%% file: main.tex
\documentclass[10pt,twocolumn,letterpaper]{article}

\usepackage{cvpr}              

\input{preamble}

\definecolor{cvprblue}{rgb}{0.21,0.49,0.74}
\usepackage[pagebackref,breaklinks,colorlinks,allcolors=cvprblue]{hyperref}

\def\confName{CVPR}
\def\confYear{2026}

 \title{Revealing Perception and Generation Dynamics in LVLMs: Mitigating Hallucinations via Validated Dominance Correction}

\author{Guangtao Lyu$^{1}$, Xinyi Cheng$^{2}$, Chenghao Xu$^{3}$, Qi Liu$^{1}$, Muli Yang$^{4}$, Fen Fang$^{4}$, \\ Huilin Chen$^{5}$, Jiexi Yan$^{2}$,  Xu Yang$^{1}$, Cheng Deng$^{1}$\thanks{Corresponding author} \\
        $^{1}$ School of Electronic Engineering, Xidian University, China, 
        $^{2}$ School of Computer Science and Technology, \\  Xidian University, China,  $^{3}$ Hohai university, China, 
        $^{4}$ Institute for Infocomm Research (I\textsuperscript{2}R), A*STAR, \\ Singapore,  
        $^{5}$School of Foreign Languages, Xidian University, China, \\
        \texttt{ \{guangtaolyu,qiliu,xinyicheng\}@stu.xidian.edu.cn,  fang fen@a-star.edu.sg} ,\\ 
        \texttt{\{jxyan1995,muliyang.xd,xuyang.xd,chdeng.xd\}@gmail.com, hlchen@xidian.edu.cn} }
        
\begin{document}
\maketitle
\input{sections/0.abs}

\input{sections/1.intro}

\input{sections/2.related}

\input{sections/3.method}

\input{sections/4.exp}

\input{sections/5.conclusion}
\clearpage
{
    \small
    \bibliographystyle{ieeenat_fullname}
    \bibliography{main}
}

\input{sections/appendix}

\end{document}

%% file: preamble.tex
\definecolor{c1}{HTML}{D4AF37}
\definecolor{c2}{HTML}{c23616}
\definecolor{c3}{HTML}{8c7ae6}
\definecolor{c4}{HTML}{465c69}
\definecolor{c5}{HTML}{98a886}
\definecolor{c6}{HTML}{C4A69D}
\definecolor{lgtblue}{rgb}{0.21,0.44,0.63}

\usepackage{epigraph} 
\usepackage{wrapfig}
\usepackage{url}
\usepackage{amsmath}  
\usepackage{amssymb}  
\usepackage{amsfonts} 

\usepackage{algorithm}
\usepackage{algorithmic}

\usepackage{bbding}
\usepackage{pifont}
\usepackage{wasysym}

\usepackage{verbatim}
\usepackage{booktabs}
\usepackage{multirow}
\usepackage{colortbl}
\usepackage{adjustbox}
\usepackage{makecell}
\usepackage{xspace}
\usepackage{mathtools}
\usepackage{graphicx}
\usepackage{enumitem}
\usepackage{textcomp}
\usepackage{pifont}
\usepackage{bm}

\usepackage[table]{xcolor}

%% file: sections/0.abs.tex
\begin{abstract}

Large Vision-Language Models (LVLMs) have shown remarkable capabilities, yet hallucinations remain a persistent challenge.  This work presents a systematic analysis of the internal evolution of visual perception and token generation in LVLMs, revealing two key patterns.  First, perception follows a three-stage GATE process: early layers perform a Global scan, intermediate layers Approach and Tighten on core content, and later layers Explore supplementary regions.  Second, generation exhibits an SAD (Subdominant Accumulation to Dominant) pattern, where hallucinated tokens arise from the repeated accumulation of subdominant tokens lacking support from attention (visual perception) or feed-forward network (internal knowledge).  Guided by these findings, we devise the VDC (Validated Dominance Correction) strategy, which detects unsupported tokens and replaces them with validated dominant ones to improve output reliability.  Extensive experiments across multiple models and benchmarks confirm that VDC substantially mitigates hallucinations.

\end{abstract}

%% file: sections/1.intro.tex
\vspace{-4pt}
\section{Introduction}\label{sec:intro}

\begin{figure}[t]
    \centering
    \includegraphics[width=\linewidth]{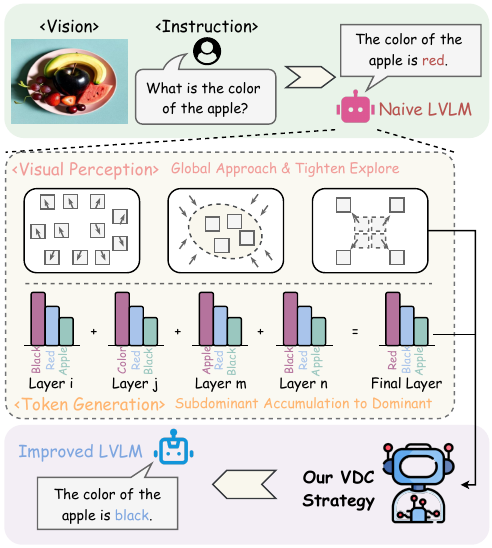}
    \vspace{-24pt}
\caption{Overview of our framework. We analyze hallucinations via internal perception and generation dynamics, revealing the \textbf{GATE (Global, Approach \& Tighten, Explore)} pattern in visual perception, the \textbf{SAD (Subdominant Accumulation to Dominant)} pattern in token generation, and devise \textbf{VDC (Validated Dominance Correction)} strategy to mitigate hallucinations.}
    \vspace{-16pt}
    \label{fig:motivation}
\end{figure}

LVLMs have achieved remarkable progress in multimodal reasoning and can handle diverse tasks~\cite{llava,gpt4,qwenvl}. However, they often hallucinate, generating content inconsistent with visual inputs or user instructions, which undermines their reliability in safety-critical domains such as medical analysis~\cite{sun2024self_medical,chen2024towards_medical,hu2024omnimedvqa_medical} and autonomous driving~\cite{jiang2024senna_driving,sun2025towards_driving,shao2024lmdrive_driving}.

The causes of hallucinations in LVLMs are complex and often attributed to multiple factors~\cite{bai2024hallucination_survey_llm,ji2023_hallucination_nlp_survey,liu2024survey_lvlm,pope}. Two aspects have received particular attention. First, models tend to overly rely on textual information, reflected in consistently lower attention to visual tokens compared to text tokens~\cite{pai_lack_visual,yue2024less_lack_visual,halc_lack_visual,favero2024multi_lack_visual,hallu_bayesian_weixin_lack_visual}. This observation has motivated methods aimed at increasing visual attention weights~\cite{pai_lack_visual,an2025mitigating_agla__lack_visual} or, conversely, reducing them as negative examples to induce hallucinations and contrast against the original input~\cite{vcd_lack_visual,second_lack_visual,opera_lack_visual}. Second, a perception–generation mismatch has been observed: even when the model correctly perceives relevant visual content, later outputs can abruptly change, resulting in incorrect tokens~\cite{skeanlayer_layer_evolution_middle_better,dola,Seeing_but_Not_Believing_1}. These findings suggest that intermediate layers may provide more reliable signals than the final layers~\cite{skeanlayer_layer_evolution_middle_better,dola_hidden_life_vista}, and some works have proposed leveraging intermediate representations to correct corresponding signals in later layers to improve reliability~\cite{dola,dola_deco,dola_looking_twice_memvr_ffn_iternal}.

Although many methods have been proposed to mitigate hallucinations, the underlying mechanisms behind these phenomena remain poorly understood. Why can a model, despite exhibiting relatively low visual attention ratio in the middle layers, still focus accurately on task-relevant regions? And why do errors sometimes emerge in the final layers, even after earlier layers have correctly attended to key visual cues~\cite{Seeing_but_Not_Believing_1,Seeing_but_Not_Believing_3,Seeing_but_Not_Believing_2,Seeing_but_Not_Believing,Seeing_but_Not_Believing_4}? These questions suggest that static observations of attention ratios or layer-wise outputs are insufficient to explain the model’s behavior. Motivated by this, we take a dynamic and evolutionary perspective, systematically examining how visual perception and token generation evolve across layers to uncover the internal processes underlying hallucination formation.

To investigate how LVLMs perceive multimodal inputs, we analyze the evolution of visual attention across layers using complementary tools. We introduce two analytical perspectives, {stage-to-global} and {inter-stage attention difference maps}, to reveal the internal attention dynamics. Our analysis uncovers a structured three-stage process, termed \textbf{GATE (Global, Approach and Tighten, Explore)}.
In the \textit{Global} stage, attention is broadly distributed across the image for holistic exploration. During the \textit{Approach and Tighten} stages, it progressively focuses on task-relevant regions and suppresses peripheral cues, often reducing overall attention ratios despite improved grounding. In the final \textit{Explore} stage, attention redistributes toward complementary regions for validation and refinement before generation.
The GATE pattern explains why intermediate layers may show lower visual attention ratios while still achieving accurate grounding, as the model tightens its focus after understanding the instruction.

We further investigate the internal evolution of token generation by adopting the {logit lens}~\cite{logits_lens} to project hidden states onto the vocabulary space. Beyond the whole-layer view, we analyze submodules, including the attention and feed-forward network (FFN), where the attention reflects \textit{visual perception}~\cite{vcd_lack_visual,pai_lack_visual} and the FFN encodes \textit{ internal parametric knowledge}~\cite{geva2021transformer_ffn_internal,dai2022knowledge_ffn_internal,jie2024memory_ffn_internal,yao2024knowledge_ffn_internal}. This intra-layer analysis reveals how perception and knowledge jointly shape token distributions.
We find that early layers exhibit token fluctuations, while middle layers gradually stabilize semantically coherent candidates, consistent with the GATE perception pattern.
We further identify the \textbf{SAD (Subdominant Accumulation to Dominant)} pattern, in which hallucinated tokens never appear as dominant (rank-1) candidates in either the attention or FFN outputs. Instead, they persist as subdominant candidates across multiple layers and progressively accumulate through cross-layer aggregation, eventually overtaking the originally correct tokens. This provides a mechanistic explanation for why LVLMs may “see correctly but speak wrongly”~\cite{Seeing_but_Not_Believing_4,Seeing_but_Not_Believing_2} and suggests a practical diagnostic cue: tokens that never dominate attention or FFN outputs are likely unreliable and can be replaced by previously validated dominant tokens.

Motivated by the GATE and SAD patterns, we introduce \textbf{Validated Dominance Correction (VDC)}, a training-free, plug-and-play strategy for detecting and mitigating hallucinations. VDC validates each generated token by checking whether it is dominant in either the attention output (visual perception) or the FFN output (internal knowledge) at any layer. Tokens that fail this validation are replaced with the most frequently dominant token across layers, ensuring that each output is supported by at least visual perception or internal knowledge. VDC effectively suppresses unreliable tokens and enhances output reliability. Extensive experiments confirm its effectiveness in reducing hallucinations and validating our internal dynamic analysis.

In summary, this work makes three key contributions:
\begin{itemize}
    \item We perform a {multi-perspective analysis within each internal LVLM layer}, jointly examining attention and FFN outputs to understand how perception and generation dynamically evolve and how hallucinations arise.
    \item We discover two patterns, {GATE} for perception and {SAD} for generation, which elucidate the evolution of visual perception and token generation, offering mechanistic insights into hallucination formation.
    \item We propose VDC, a training-free, plug-and-play method that uses validated dominant token to correct invalidated one, mitigating hallucinations across benchmarks.
\end{itemize}

%% file: sections/2.related.tex
\section{Related Work}\label{sec:related-work}
\vspace{-4pt}

\noindent\textbf{LVLMs.}
The success of LLMs~\cite{gpt1,llama,vicuna,deepseek,qwenllm,deepseekr1} has enabled LVLMs~\cite{instructblip,llava,qwenvl,anthropic2024claude,gpt4,mplugowl2,minigpt4,hu2024minicpm,llavanext}, which integrate visual and textual information for multimodal understanding and reasoning. Typical LVLMs use a pre-trained visual encoder to extract image features, projected into the LLM embedding space via linear layers or Q-Former modules~\cite{llava1.5,chen2024internvl,yang2025qwen3}, and then combined with textual inputs. LVLMs achieve strong performance in various tasks~\cite{jiang2024senna_driving,hu2024omnimedvqa_medical,minigpt4,tmr_guangtao_1,chenghao_2,mplugowl2,lin2024videollava}. Despite these advances, hallucination remains a major challenge~\cite{lee2018hallucinations,gunjal2024detecting,liu2024survey_lvlm,woo2025miss_forest_tree_attn_vision_calibration}.

\begin{figure*}[t]
    \centering
    \includegraphics[width=0.99\linewidth]{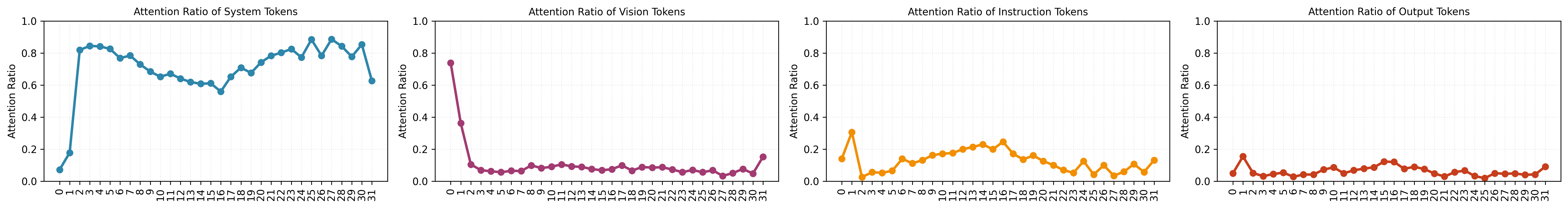}
    \vspace{-10pt}
    \caption{Attention ratio of different token types (System, Vision, Instruction, Output) across layers.}
    \vspace{-10pt}
    \label{fig:attn_ratio_layer}
\end{figure*}

\begin{figure*}[t]
    \centering
    \includegraphics[width=0.99\linewidth]{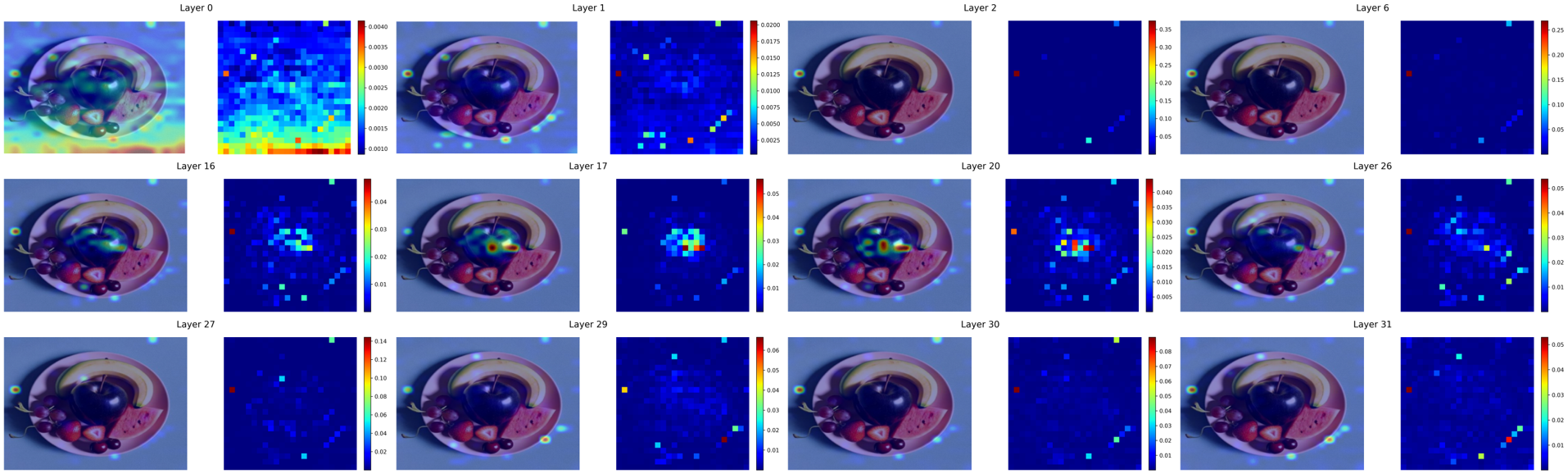}
    \vspace{-10pt}
    \caption{Visual attention heatmaps across layers.}
    \vspace{-15pt}
    \label{fig:attn_heatmap_layer}
\end{figure*}

\begin{figure}[t]
    \centering
    \includegraphics[width=0.99\linewidth]{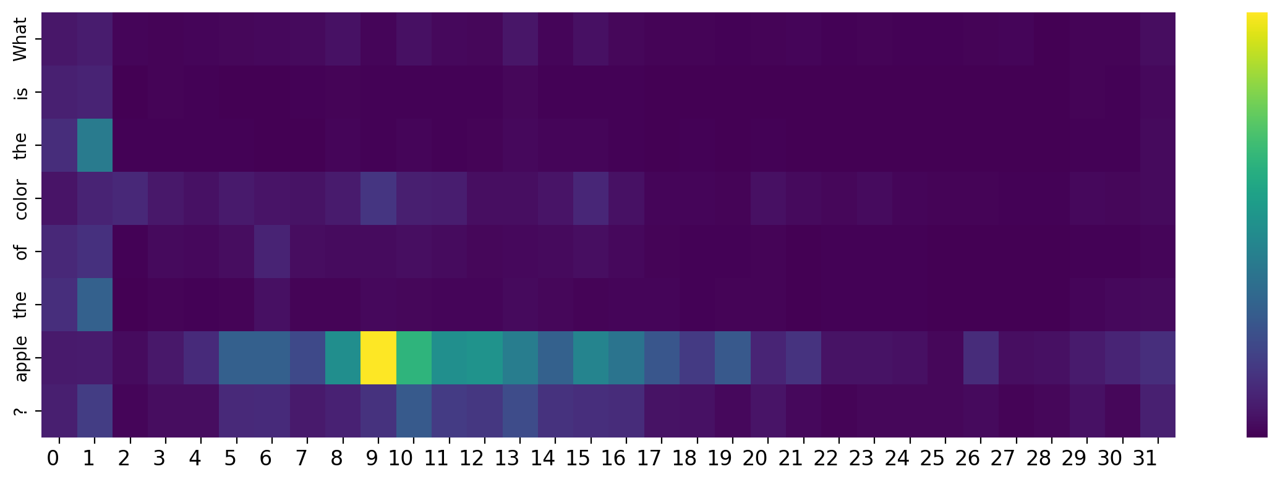}
    \vspace{-10pt}
    \caption{Instruction attention heatmaps across layers.}
    \label{fig:text_attn_layer_single}
\end{figure}

\noindent\textbf{Hallucinations in LVLMs.}
Hallucinations refer to content that is irrelevant, factually incorrect, or inconsistent with visual inputs~\cite{liu2024survey_lvlm,yue2024less_lack_visual,hallu_attention_lens}. Their emergence in LVLMs is complex, influenced by multiple factors~\cite{liu2024survey_lvlm,bai2024hallucination_survey_llm}. A common issue is over reliance on textual information~\cite{m3id_lack_visual,pai_lack_visual}, often reflected in lower attention to visual tokens; prior works address this by increasing visual attention~\cite{pai_lack_visual,second_lack_visual,opera_lack_visual,xie2025tarac_lack_visual,huo2024self_sid_lack_visual} or using reduced visual attention as negative examples~\cite{vcd_lack_visual,m3id_lack_visual,an2025mitigating_agla__lack_visual,jianginterpreting_hallu_logits_lens}. Another key observation is the perception–generation mismatch: even when visual content is correctly perceived, outputs may produce incorrect tokens~\cite{skeanlayer_layer_evolution_middle_better,Seeing_but_Not_Believing_1,Seeing_but_Not_Believing_2,Seeing_but_Not_Believing_3}. Some studies leverage middle layers to improve final predictions~\cite{dola,dola_deco,dola_hidden_life_vista} or selectively skip late layers~\cite{elbayad2019depth_skip_layers,zhang2025redundancy_evolution_information_flow,dola_looking_twice_memvr_ffn_iternal}. Despite these efforts, understanding remains limited. We study hallucinations through internal perception and generation dynamics, identifying the {GATE} pattern in visual perception, the {SAD} pattern in token generation, and proposing {VDC} to mitigate hallucinations.

%% file: sections/3.method.tex
\begin{figure*}[t]
    \centering
    \includegraphics[width=0.99\linewidth]{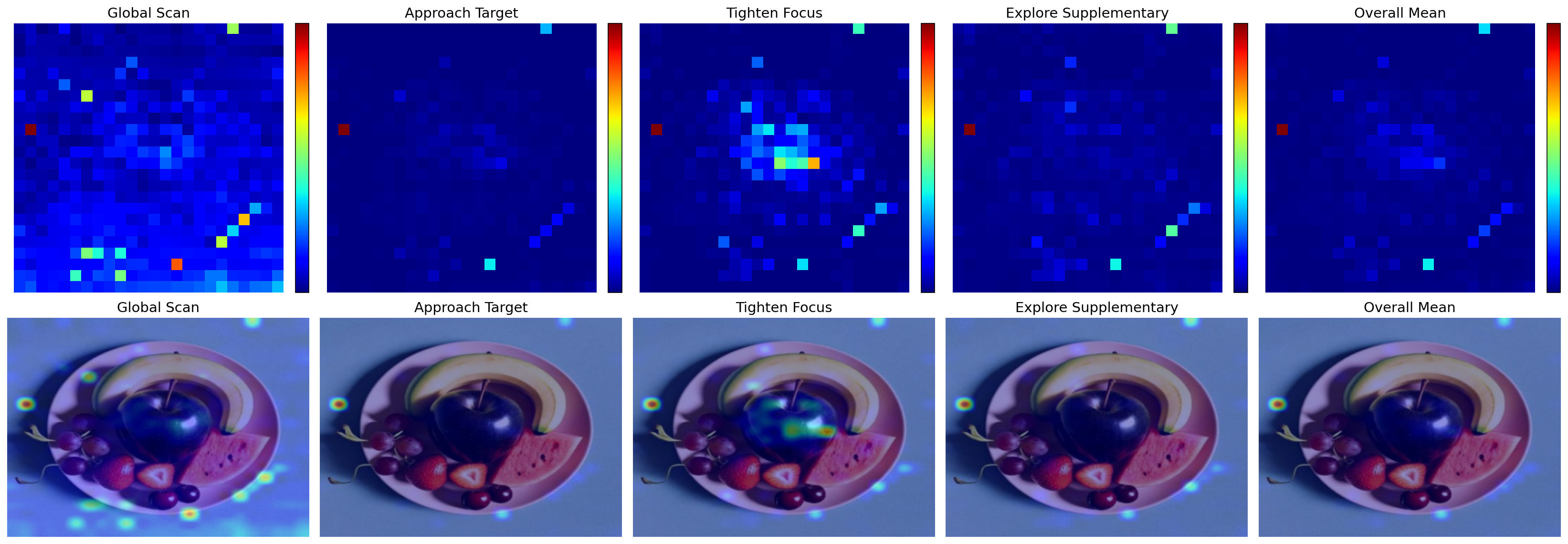}
    \vspace{-10pt}
\caption{
Heatmaps of visual attention across different stages, illustrating the GATE pattern (Global–Approach\&Tighten–Explore). 
In the Global stage, the model attends broadly to the entire image; in the Approach phase, attention gradually shifts toward the apple region; during the Tighten phase, focus converges tightly on the apple; and in the final Explore stage, attention expands again to nearby areas.
}
    \vspace{-6pt}
    \label{fig:attn_stages}
\end{figure*}

\begin{figure*}[t]
\begin{minipage}{0.49\textwidth}
    \centering
    \includegraphics[width=\linewidth]{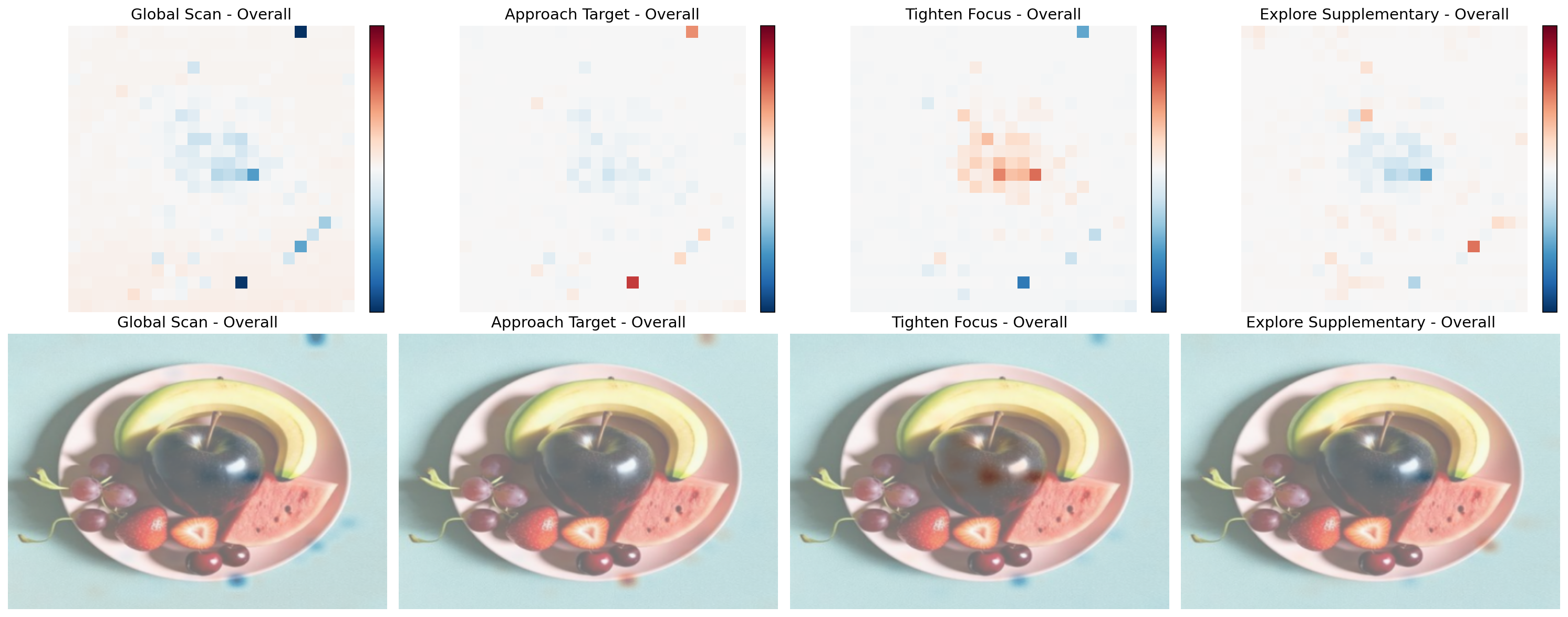}
    \vspace{-20pt}
    \caption{Stage-to-global attention difference maps, showing how each stage's attention deviates from the overall average, highlighting stage-specific focus or deficiencies.}
    \vspace{-10pt}
    \label{fig:attn_differ_overall}
\end{minipage}
\hfill
\begin{minipage}{0.49\textwidth}
    \centering
    \includegraphics[width=0.75\linewidth]{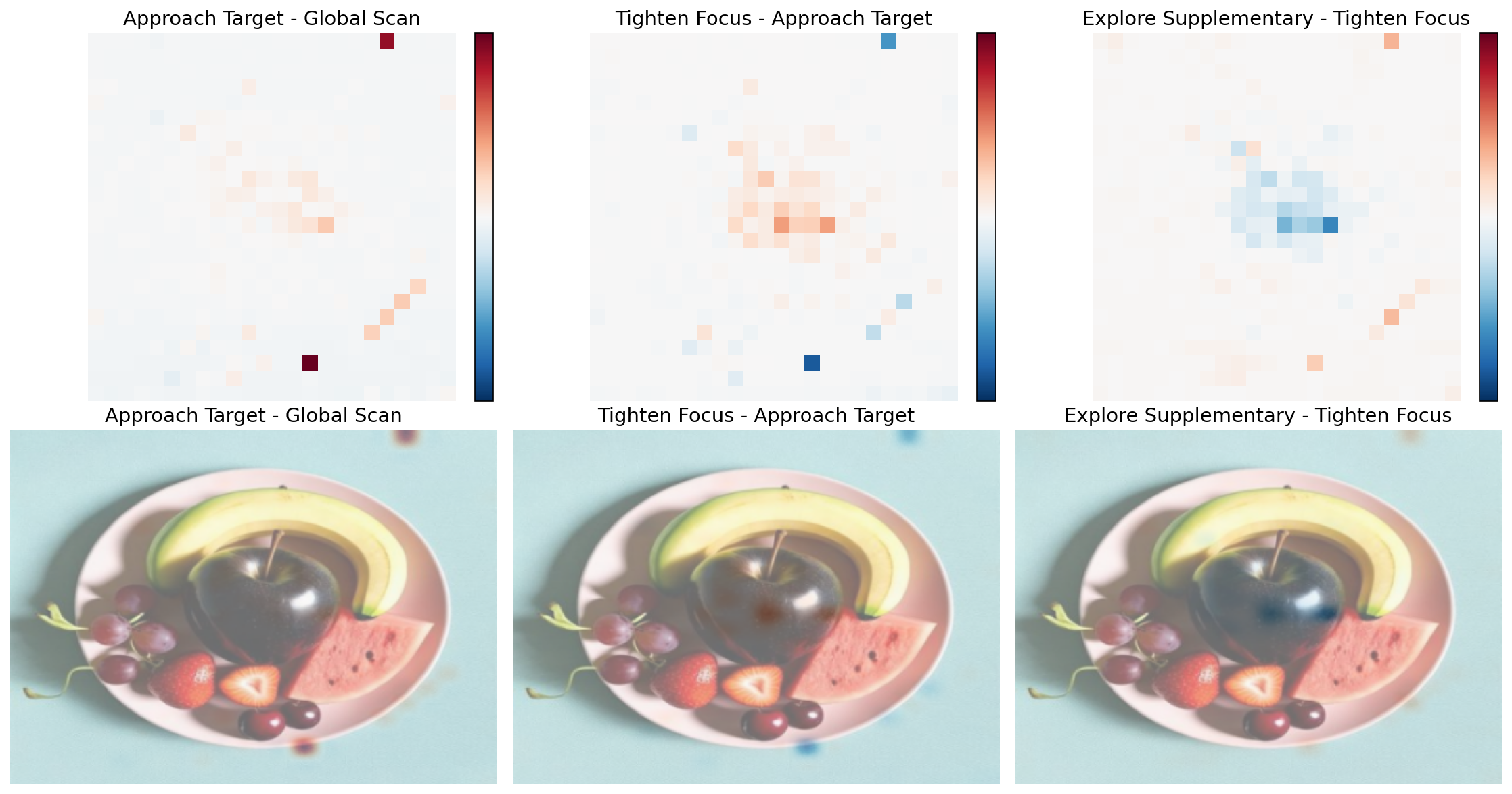}
    \vspace{-10pt}
    \caption{Inter-stage attention difference maps, showing how attention shifts across consecutive stages, revealing the model's dynamic focus evolution and the sequence of information integration.}
    \vspace{-10pt}
    \label{fig:attn_differ_inter}
\end{minipage}
\end{figure*}

\begin{figure}[t]
    \centering
    \includegraphics[width=0.92\linewidth]{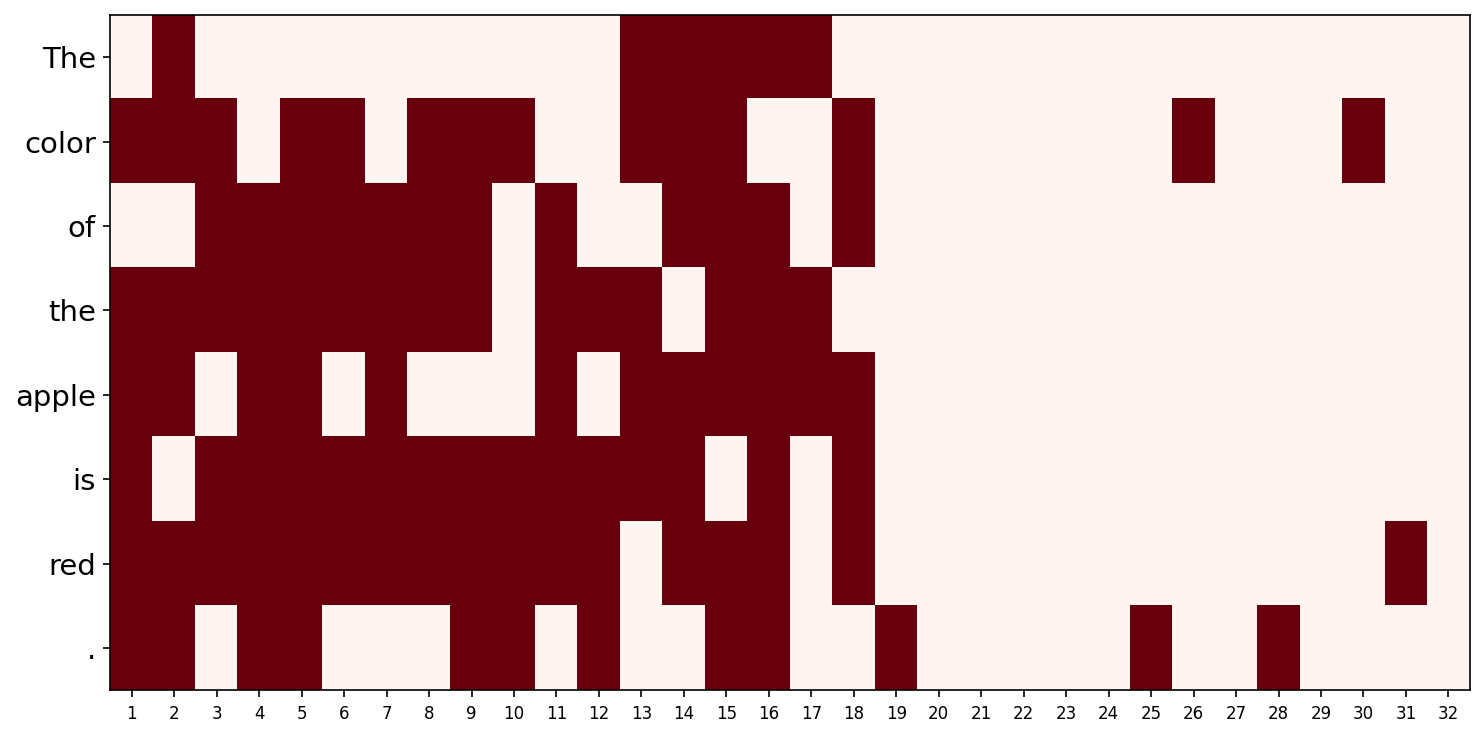}
    \vspace{-10pt}
\caption{Dominant (Top-1) token changes across layers. Red means changes from the previous layer, white means no change.}
    \vspace{-10pt}
    \label{fig:token_change_rank1}
\end{figure}

\begin{figure*}[t]
    \begin{minipage}{\textwidth}
        \centering
        \includegraphics[width=\linewidth]{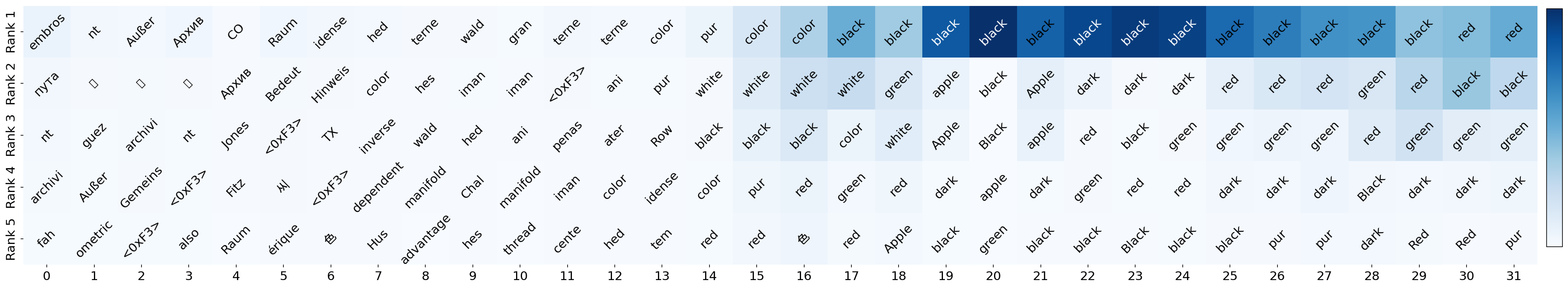}
        \vspace{-20pt}
\caption{Top-5 tokens from layer outputs. Early layers exhibit rapid token fluctuations, followed by a stable phase producing the correct token ``black''. The last two layers show a sudden shift to the incorrect token ``red''.}
        \label{fig:token_layer_out}
    \end{minipage}
    
    \begin{minipage}{\textwidth}
        \centering
        \includegraphics[width=\linewidth]{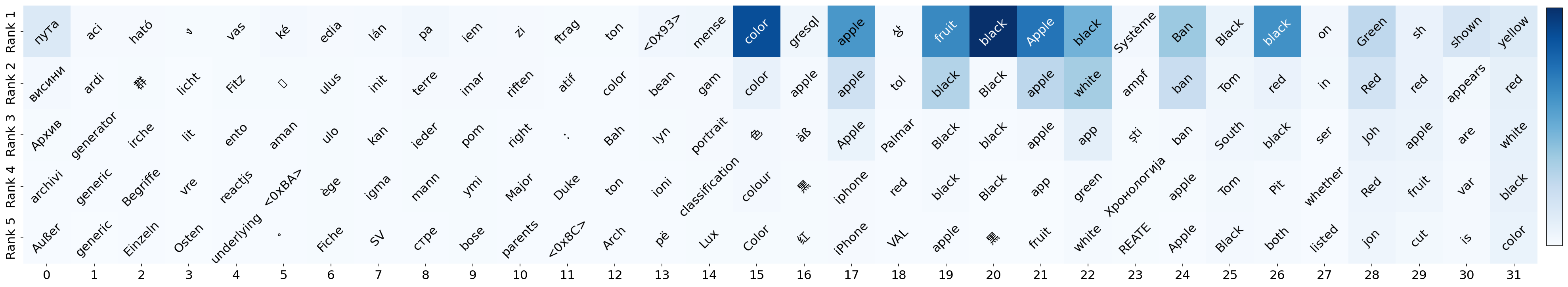}
        \vspace{-20pt}
\caption{Top-5 tokens from attention outputs. The token ``red'' does not appear as the dominant (rank-1) token in any layer, indicating that it fails to receive support or validation from visual perception.}
        \label{fig:token_attn_out}
    \end{minipage}

    \begin{minipage}{\textwidth}
        \centering
        \includegraphics[width=\linewidth]{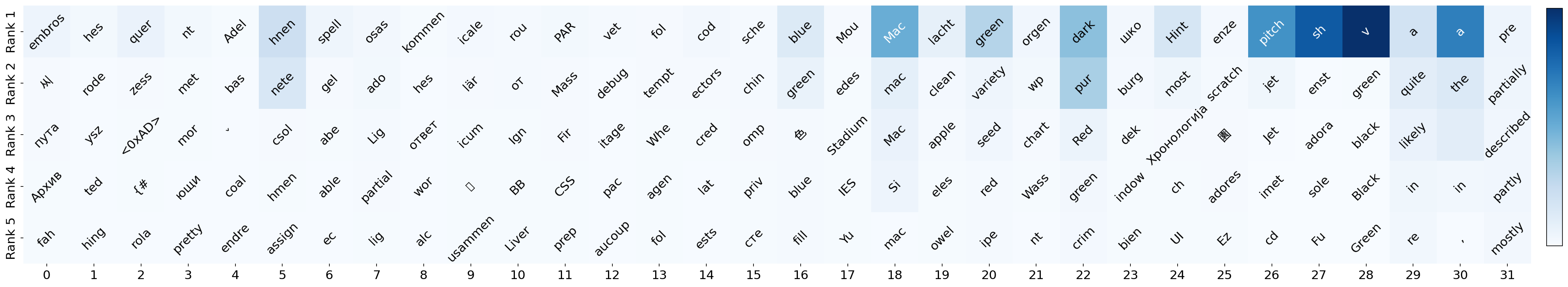}
        \vspace{-20pt}
\caption{Top-5 tokens from FFN outputs. The token ``red'' also does not appear as the dominant (rank-1) token in any layer, indicating that it also fails to receive support or validation from the model's internal parametric knowledge.}
        \label{fig:token_ffn_out}
    \end{minipage}
\end{figure*}

\section{Analysis of LVLM Internal Dynamics}
\noindent\textbf{Preliminaries.}
We denote the system message tokens, visual tokens, instruction tokens, and output tokens as $\mathbf{X}_s$, $\mathbf{X}_v$, $\mathbf{X}_i$, and $\mathbf{X}_o$, respectively~\cite{llava}.
Each LVLM layer~\cite{transformer} consists of an attention module, an FFN, and normalization modules. 
We denote the hidden states of the attention module, FFN, and the layer as $h^{\text{attn}}$, $h^{\text{ffn}}$, and $h$:
\begin{align}
\small
h^{\text{attn}} &= \text{Attention}(\text{Norm}(h)), \\
h &= h + h^{\text{attn}}, \\
h^{\text{ffn}} &= \text{FFN}(\text{Norm}(h)), \\
h &= h + h^{\text{ffn}}.
\end{align}
We apply \textit{logits lens}~\cite{logits_lens} to project them into the language space for an internal examination of token generation.

\subsection{Perception: The GATE Pattern}

\noindent\textbf{Layer-Wise Attention Dynamics.}
To understand how LVLMs perceive and integrate multimodal information,
we analyze the evolution of visual and textual attention across layers using both qualitative and quantitative approaches.
Specifically, we visualize the layer-wise attention ratio across four token groups: system, vision, instruction, and output (Fig.~\ref{fig:attn_ratio_layer}),
as well as the attention from generation tokens to visual and instruction tokens (Figs.~\ref{fig:attn_heatmap_layer},~\ref{fig:text_attn_layer_single}).

From Figs.~\ref{fig:attn_ratio_layer},~\ref{fig:attn_heatmap_layer},~\ref{fig:text_attn_layer_single}, we observe changes in both attention spatial focus and allocation across layers. 
In the first two layers, attention over visual tokens is widespread, covering nearly the entire image, indicating a global scanning process.
At layer~3, attention to system tokens increases sharply while attention to other token types decreases.
Between layers~3 and~16, both visual and instruction attentions increase steadily, as the model understands key instruction cues and focuses on the corresponding visual regions.
Attention to the instruction peaks around layer~16, while at layer~17 visual attention concentrates on the main object region.
From layers~17 to~26, attention tightens on key object regions while peripheral regions receive less focus, slightly lowering overall visual and instruction ratios.
At layer~27, attention in the key region begins to disperse, shifting toward complementary regions. In the final layer, visual and instruction attentions show a rebound.

\noindent\textbf{Stage-Level Interpretation and Visualization.}
Building on the layer-wise attention dynamics, we summarize the evolving perception process into three stages, forming the GATE pattern (Global, Approach \& Tighten, Explore), which reflects systematic shifts in spatial focus and task-relevant allocation across layers.
To illustrate these stages, we visualize representative stage-specific attention maps (Fig.~\ref{fig:attn_stages}), compute stage-to-global differences to highlight emphasized or suppressed regions (Fig.~\ref{fig:attn_differ_overall}), and analyze inter-stage differences to capture the transitions between consecutive stages (Fig.~\ref{fig:attn_differ_inter}). 
Such differential visualizations also help mitigate the effect of attention sinks~\cite{xiao2023efficient_attention_sinks_ori,sun2024massive_activations_attention_sinks,kang2025see_attention_sinks,schuster2022confident_skip_layers}, providing a clearer and more interpretable depiction of stage-wise evolution.
Additional examples are provided in Appendix Figs.~\ref{fig:example_black_apple},~\ref{fig:example_right_brick},~\ref{fig:example_dog_run_away},~\ref{fig:example_person_sitting},~\ref{fig:example_dog_standing_walk},~\ref{fig:example_woman_behind_ahead},~\ref{fig:example_ffn_right_standing}, and~\ref{fig:examples_right_balck_apple}.

In the \textit{Global Scanning} stage, attention is broadly distributed across the image, characterized by high visual ratios. Stage-to-global difference maps highlight stronger attention in surrounding regions (red) and weaker attention at the center (blue), reflecting a holistic scanning process.
In the \textit{Approach \& Tighten} stage, the model begins integrating textual guidance and gradually {approaches} task-relevant regions, directing increased attention to key objects, such as the apple, as shown by the difference maps (red). It then {tightens} its focus on the most informative object and instruction tokens, while suppressing peripheral areas (center red, surroundings blue), demonstrating refined attention to critical cues.
In the \textit{Explore Supplementary} stage, attention expands to peripheral or complementary regions. Difference maps reveal reduced focus on previously attended key regions (blue) alongside increased attention to secondary areas (red), reflecting a rechecking process that reassesses surrounding regions while retaining critical information.

\noindent\textbf{The GATE Pattern.}
The GATE characterizes how attention is dynamically allocated and reallocated across layers in LVLMs.
However, even when the model correctly attends to core visual regions, it can still generate incorrect outputs, revealing a perception–generation mismatch~\cite{Seeing_but_Not_Believing_4,Seeing_but_Not_Believing_2}.
To uncover the source of this discrepancy, we next analyze the token generation process in detail.

\subsection{Generation: The SAD Pattern}

\noindent\textbf{Layer-Wise Generation Analysis.}
As shown by the GATE analysis, even when LVLMs achieve correct visual grounding, they can still produce hallucinated outputs. To investigate this perception–generation mismatch, we perform a layer-wise analysis of token predictions using the \textit{logit-lens}~\cite{logits_lens}, projecting hidden states into the vocabulary space (Fig~\ref{fig:token_layer_out}). We also track whether the rank-1 (dominant) token changes across layers (Figs~\ref{fig:token_change_rank1}), standardizing all tokens to lowercase to ignore superficial syntactic differences and focus on semantic changes (see Appendix Figs.~\ref{fig:example_black_apple},~\ref{fig:example_right_brick},~\ref{fig:example_dog_run_away},~\ref{fig:example_person_sitting},~\ref{fig:example_dog_standing_walk},~\ref{fig:example_woman_behind_ahead},~\ref{fig:example_ffn_right_standing}, and~\ref{fig:examples_right_balck_apple} for more examples).

In Fig.~\ref{fig:token_layer_out}, during the early layers, corresponding to the Global and Approach stages of the GATE, predicted tokens are largely unrelated to the target, gradually transitioning from random content words to task-relevant terms, such as color attributes. Around layer 17, the model begins to predict the correct token “black,” coinciding with the peak instruction attention (Fig.~\ref{fig:attn_ratio_layer}) and the visual focus on the “black apple” region (Fig.~\ref{fig:attn_heatmap_layer}). The token “black” remains dominant throughout the middle layers, covering the entire Tighten stage, until the final two layers, where it abruptly shifts to the incorrect token “red”.

In Fig.~\ref{fig:token_change_rank1}, early layers exhibit frequent token fluctuations, corresponding roughly to the Global Scanning and early Approach phases, during which the model broadly explores the image and interprets the user instruction. By the middle layers, simple structural tokens such as ``of", ``apple", ``object", and ``image'' begin to stabilize, indicating that basic and simple words are resolved relatively early. In contrast, content-rich tokens such as ``red", ``brick", ``side", and ``walk'' continue to fluctuate, reflecting higher uncertainty and a greater need for semantic reasoning.

These observations indicate that intermediate layers already stabilize low-level or simple tokens. Tokens that consistently appear as the dominant prediction across multiple layers can be considered resolved, suggesting potential computational efficiency gains by skipping these tokens in subsequent processing~\cite{kao2020bert_skip_layers,luo2025adaptive_skip_layers,elbayad2019depth_skip_layers,schuster2022confident_skip_layers}.
In contrast, tokens that continue to fluctuate across layers likely reflect higher uncertainty and a greater need for semantic reasoning. This variability provides an informative signal for identifying challenging samples and can guide dataset curation or selection, focusing training or evaluation on instances where persistent token changes occur~\cite{wang2024greats_data_curation_select,bansal2025honeybee_data_curation_select,brown2025benchmark_data_curation_select}.  

 In the final two layers, token predictions often change. For example, a token like ``black''  replaced with ``red'', or ``side'' become ``brick''.
These late-stage shifts occur in both correct and hallucinated outputs, indicating that such sudden changes are an inherent property of LVLMs.

\noindent\textbf{Disentangling Intra-Layer Generation Dynamics.}
To understand why these abrupt changes occur, we further disentangle the contributions of attention and FFN within each layer. By examining the projected outputs $\mathbf{X}_o^{\text{attn}}$ and $\mathbf{X}_o^{\text{ffn}}$ separately, we can trace how visually grounded perception and internal knowledge independently influence token generation, and how their interactions lead to sudden token shifts in the final output.

Figs.~\ref{fig:token_attn_out},~\ref{fig:token_ffn_out}, and~\ref{fig:token_layer_out} show that $\mathbf{X}_o^{\text{attn}}$ and $\mathbf{X}_o^{\text{ffn}}$ changes rapidly across layers, while the whole layer output $\mathbf{X}_o$ remains comparatively stable. The final hallucinated token ``red" never appears as dominant in either branch, instead remaining subdominant tokens that gradually overtakes the correct token in $\mathbf{X}_o$ via multi-layer accumulation.

Across many examples (Figs.~\ref{fig:example_black_apple},~\ref{fig:example_right_brick},~\ref{fig:example_dog_run_away},~\ref{fig:example_person_sitting},~\ref{fig:example_dog_standing_walk},~\ref{fig:example_woman_behind_ahead},~\ref{fig:example_ffn_right_standing}, and~\ref{fig:examples_right_balck_apple} in Appendix), we observe that late-stage token shifts are common. Correct tokens that have previously appeared as dominant in attention or FFN outputs, such as ``brick," typically remain grounded, while hallucinated tokens that never dominate can gradually accumulate across layers and replace correct tokens in the final output.

\noindent\textbf{The SAD Pattern.}
We term this the \textbf{SAD} (\textit{Subdominant Accumulation to Dominant}). It describes how hallucinated tokens, which never become the dominant token in either the attention or FFN outputs, remain consistently subdominant across layers yet gradually accumulate, eventually overtaking the correct token in the final output. This provides a mechanistic explanation for why LVLMs may “see correctly but speak wrongly” and suggests a practical diagnostic cue: tokens that never dominate in  either the attention output (visual perception) or the FFN output (internal knowledge) are likely unreliable and can be replaced by previously validated dominant tokens.

\begin{algorithm}[t]
\caption{Validated Dominance Correction (VDC)}
\label{alg:vdc}
\begin{algorithmic}[1]
\REQUIRE Dominant tokens $d^{\text{attn},\ell}_t, d^{\text{ffn},\ell}_t$; output $\mathbf{X}_o$
\ENSURE Corrected output $\hat{\mathbf{X}}_o$
\STATE Initialize $\hat{\mathbf{X}}_o \gets \{\}$
\FOR{$t = 1$ to $T$}
    \STATE Generate token $\mathbf{x}_t$
    \STATE Collect dominant tokens $\{d^{\text{attn},\ell}_t, d^{\text{ffn},\ell}_t\}_{\ell=1}^{L}$
    \IF{$\mathbf{x}_t \notin \{d^{\text{attn},\ell}_t, d^{\text{ffn},\ell}_t \ \forall \ell\}$}
        \STATE Mark $\mathbf{x}_t$ as hallucinated
        \STATE Replace $\mathbf{x}_t \leftarrow \mathbf{x}^*_t$, the most frequent dominant token across layers
    \ENDIF
    \STATE Append $\mathbf{x}_t$ to $\hat{\mathbf{X}}_o$
\ENDFOR
\RETURN $\hat{\mathbf{X}}_o$
\end{algorithmic}
\end{algorithm}

\section{Validated Dominance Correction}

Motivated by the SAD and GATE patterns, we propose \textbf{Validated Dominance Correction (VDC)}, a training-free plug-and-play strategy to detecting and mitigate hallucination. In essence, if a generated token never appears as the dominant token in either the attention (visual perception) or FFN (internal knowledge) outputs, it is considered unreliable and treated as a hallucination. Such tokens are immediately replaced with a validated token that has been dominant in at least one output within the current generation step, ensuring that every token in the output is grounded in either visual perception or knowledge-based reasoning.

Let $\mathbf{X}_o = \{\mathbf{x}_1, \dots, \mathbf{x}_T\}$ denote the sequence of generated tokens.  
At each generation step $t$, we obtain layer-wise dominant tokens from attention and FFN modules, denoted by $d^{\text{attn},\ell}_t$ and $d^{\text{ffn},\ell}_t$ for $\ell = 1, \dots, L$.  
A generated token $\mathbf{x}_t$ is considered \textit{validated} if it appears as the dominant token in either output at any layer:
\begin{equation}
\small
    \text{validated}(\mathbf{x}_t) = 
\begin{cases}
1, & \exists \ell \text{ s.t. } \mathbf{x}_t = d^{\text{attn},\ell}_t \text{ or } \mathbf{x}_t = d^{\text{ffn},\ell}_t \\
0, & \text{otherwise}.
\end{cases}
\end{equation}

If $\text{validated}(\mathbf{x}_t) = 0$, the token is treated as a hallucination and replaced by the most frequent dominant token $\mathbf{x}^*_t$ across layers:
\begin{equation}
\small
\mathbf{x}^*_t = \arg\max_{\mathbf{x} \in \{d^{\text{attn},\ell}_t, d^{\text{ffn},\ell}_t\}} 
\sum_{\ell=1}^{L} 
\mathbb{I}[\mathbf{x} = d^{\text{attn},\ell}_t \text{ or } \mathbf{x} = d^{\text{ffn},\ell}_t].
\end{equation}

The corrected output is updated online as:
\begin{equation}
\small
\hat{\mathbf{x}}_t =
\begin{cases}
\mathbf{x}_t, & \text{if } \text{validated}(\mathbf{x}_t) = 1 \\
\mathbf{x}^*_t, & \text{if } \text{validated}(\mathbf{x}_t) = 0
\end{cases}, \quad t = 1,\dots,T.
\end{equation}

%% file: sections/4.exp.tex
\input{tables/pope}

\input{tables/chair}

\section{Experiments}
\label{sec:experiment}

\textbf{Benchmarks.}  
We evaluate on three standard hallucination benchmarks:
(1) \textbf{POPE}~\cite{pope}, which measures object-level hallucinations through binary yes/no questions about object existence.
(2) \textbf{CHAIR}~\cite{chair}, which assesses hallucinations in open-ended image captioning on 500 randomly selected images from the MSCOCO~\cite{mscoco} validation set. The metric is mainly reported at both instance-level ($\text{CHAIR}_\text{I}$) and sentence-level ($\text{CHAIR}_\text{S}$):
$
\text{CHAIR}_I = \frac{|\{\text{hallucinated objects}\}|}{|\{\text{all objects}\}|}, \quad
\text{CHAIR}_S = \frac{|\{\text{captions containing hallucinations}\}|}{|\{\text{captions}\}|}.
$
(3) \textbf{MME}~\cite{mme}, which provides a comprehensive evaluation across four subsets—\textit{existence}, \textit{count}, \textit{position}, and \textit{color}—covering both object-level and attribute-level hallucinations.

\noindent \textbf{Evaluated LVLMs.}  
Following prior works~\cite{vcd_lack_visual,only_lack_visual}, we evaluate VDC on three open-source LVLMs: \textbf{LLaVA-1.5}~\cite{llava}, \textbf{InstructBLIP}~\cite{instructblip}, and \textbf{Qwen-VL}~\cite{qwenvl}. Unless specified, we use \textbf{LLaVA-1.5} as the default model.  

\noindent \textbf{Baselines.}  
We compare VDC with the following approaches: 
Vanilla, VCD~\cite{vcd_lack_visual}, M3ID~\cite{m3id_lack_visual}, and ONLY~\cite{only_lack_visual}.  
Vanilla denotes the standard LVLM decoding strategy, where the next token is directly sampled from the post-softmax probability distribution following prior works~\cite{only_lack_visual,vcd_lack_visual}.  
VCD contrasts the logits between the original and noisy images.  
M3ID contrasts the logits from inputs with and without visual information.  
ONLY leverages the text-to-vision entropy ratio of each word to selectively amplify visually grounded textual cues.  \textbf{More LVLMs and Baselines are shown in Appendix.~\ref{sec:app_more_models_llavanext_chair}}

\noindent \textbf{Implementation Details.}  
Our VDC is a \textit{plug-and-play} strategy that integrates seamlessly into existing LVLMs and  mitigation strategies, requiring only validation and correction of final tokens.  
Unlike contrastive decoding methods needing multiple forward passes, VDC collects multi-layer logits in a single pass and performs VDC through efficient tensor operations.  
It thus achieves near inference speed to vanilla decoding with consistently improved reliability.  
All experiments are conducted on a single NVIDIA RTX A6000 GPU (48GB).

\subsection{Main Experimental Results}

\noindent \textbf{Results on POPE.}  
As shown in Tab.~\ref{tab:POPE}, integrating our VDC into different baselines consistently improves performance across various LVLM backbones and evaluation settings, demonstrating its robustness and general applicability.

\noindent \textbf{Results on MME.}  
As shown in Tab.~\ref{tab:MME}, incorporating VDC into existing baselines consistently improves both object-level metrics (Existence, Count) and attribute-level metrics (Position, Color), indicating that VDC enhances the model’s overall multimodal understanding.

\begin{figure}[t]
    \centering
    \includegraphics[width=0.95\linewidth]{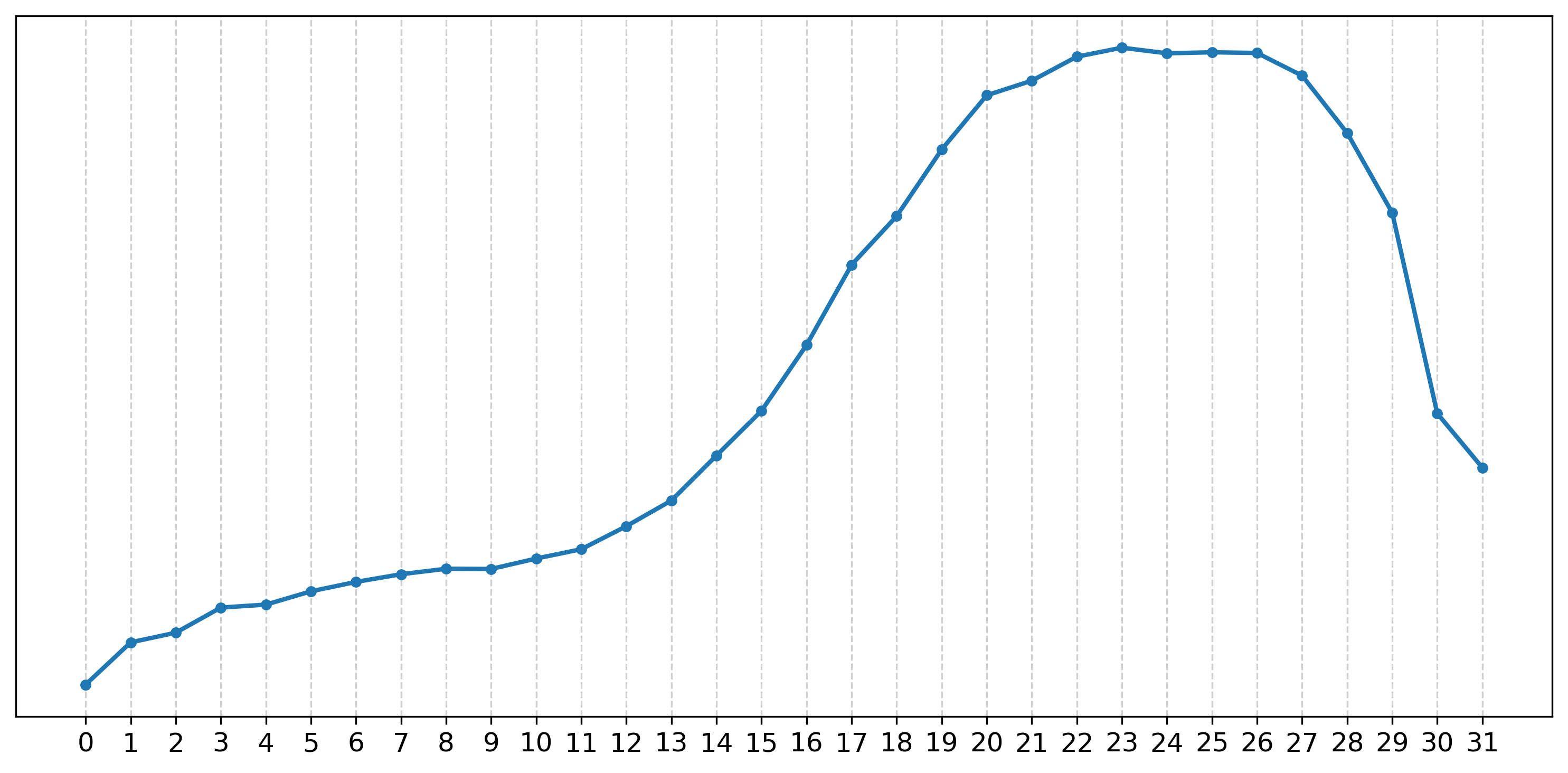}
    \vspace{-10pt}
\caption{Frequency distribution of layers where the validated correction tokens appear during VDC on CHAIR Benchmark.}
\vspace{-10pt}
    \label{fig:token_replace_frequency}
\end{figure}

\noindent \textbf{Results on CHAIR.}  
On the open-ended CHAIR benchmark (Tab.~\ref{tab:CHAIR}), integrating VDC into different models significantly reduces hallucination rates across all settings. The improvement is particularly pronounced for longer generation sequences (128 tokens), where hallucinations tend to accumulate. This demonstrates that identifying unsupported tokens and replacing them with validated ones yields more faithful and visually grounded outputs. Fig.~\ref{fig:token_replace_frequency} shows the layer-wise frequency of tokens used for correction. While some correction tokens appear across multiple layers, their occurrences are notably concentrated after layer 17, corresponding to the \textit{Tighten} phase identified by the GATE analysis. These results further validate our analysis of the internal perception and generation dynamics of LVLM.

\input{tables/mme}

\subsection{Ablation Studies}

\textbf{Validation Sources.}  
We evaluate the impact of different validation sources by using dominant tokens from {Layer}-only, {Attn+FFN}, and {Attn+FFN+Layer}. As shown in Tab.~\ref{tab:vdc_ablation_validation}, {Attn+FFN} consistently achieves the lowest hallucination rates, whereas including layer outputs ({Attn+FFN+Layer}) increases the hallucination rate. This increase can be attributed to the \emph{SAD pattern}, where subdominant incorrect tokens gradually accumulate to dominate, causing some hallucinated tokens to be mistakenly considered validated and thereby amplifying hallucinations.  In contrast, using {Attn+FFN} means a token needs to appear in either visual perception or internal knowledge to be considered validated, providing a more precise criterion for reliability.

\noindent \textbf{Correction Sources.}  
We evaluate the impact of different sources on VDC performance by testing corrections using dominant tokens from {Layer}-only, {Attn+FFN}, and {Attn+FFN+Layer}. As shown in Tab.~\ref{tab:vdc_ablation_replacement}, using {Attn+FFN} for replacement achieves the lowest hallucination rates, indicating that cross-stream attention and FFN signals provide the most precise corrections. Including the whole layer output ({Attn+FFN+Layer}) increases recall but slightly raises hallucination rates, suggesting that layer outputs help coverage at the expense of precision. To balance correction accuracy and coverage, we adopt {Attn+FFN+Layer} as the default source for VDC.

\noindent \textbf{Skipping Early Layers.}
We evaluate whether early layers contribute meaningfully to VDC. To this end, we skip the first 2, 10, and 16 layers in both validation and correction sources. As shown in Tab.~\ref{tab:skip_layers_ablation}, skipping these layers has negligible impact on hallucination rates. Analysis of perception and generation dynamics reveals that early layers mainly produce unstable or semantically uninformative tokens, consistent with the \textit{Global} and \textit{Approach} phases in the GATE pattern, during which the model is still interpreting the instruction and gradually attending to relevant regions.  These findings suggest that the first ten layers can be safely excluded, improving efficiency while confirming that the GATE pattern effectively guides VDC optimization.

\input{tables/ablation}

%% file: tables/pope.tex
\begin{table*}[t]
    \vspace{-10pt}
    \caption{Results on the POPE. $\uparrow$ indicates higher is better. The best and second results are \textbf{bolded} and \underline{underlined}.}
    \vspace{-10pt}
    \label{tab:POPE}
    
    \centering
    \small
    \setlength{\tabcolsep}{5pt}
    \resizebox{\textwidth}{!}{
    \begin{tabular}{cclcccccccccccc}
    \toprule
     & \multirow{2}{*}{{Setup}} & \multirow{2}{*}[-2pt]{{Method}} & \multicolumn{4}{c}{{LLaVA-1.5}} & \multicolumn{4}{c}{{InstructBLIP}} & \multicolumn{4}{c}{{Qwen-VL}} \\
    \arrayrulecolor{gray} \cmidrule(lr){4-7} \cmidrule(lr){8-11} \cmidrule(lr){12-15}
     &  &  & {Acc.} $\uparrow$ & {Prec.} $\uparrow$ & {Rec.} $\uparrow$ & {F1} $\uparrow$ & {Acc.} $\uparrow$ & {Prec.} $\uparrow$ & {Rec.} $\uparrow$ & {F1} $\uparrow$ & {Acc.} $\uparrow$ & {Prec.} $\uparrow$ & {Rec.} $\uparrow$ & {F1} $\uparrow$ \\
    \midrule
\multirow{24}{*} & \multirow{8}{*}{Random}
 &  Vanilla  & 84.63 & 83.07 & 87.00 & 84.99 & 83.33 & 82.38 & 84.80 & 83.57 & 85.17 & 97.22 & 72.40 & 83.00 \\
 &  & Vanilla + VDC  
 & \cellcolor{cvprblue!15} 84.90 & \cellcolor{cvprblue!15} 83.24 & \cellcolor{cvprblue!15} 87.40 & \cellcolor{cvprblue!15} 85.27 
 & \cellcolor{cvprblue!15} 83.70 & \cellcolor{cvprblue!15} 82.68 & \cellcolor{cvprblue!15} 85.27 & \cellcolor{cvprblue!15} 83.95 
 & \cellcolor{cvprblue!15} 85.73 & \cellcolor{cvprblue!15} 97.35 & \cellcolor{cvprblue!15} 73.47 & \cellcolor{cvprblue!15} 83.74 \\
 &  & VCD  & 84.57 & 82.59 & 87.60 & 85.02 & 84.60 & 85.12 & 83.87 & 84.49 & 87.37 & 97.14 & 77.00 & 85.91 \\
 &  & VCD + VDC  
 & \cellcolor{cvprblue!15} 84.93 & \cellcolor{cvprblue!15} 83.00 & \cellcolor{cvprblue!15} 87.87 & \cellcolor{cvprblue!15} 85.36 
 & \cellcolor{cvprblue!15} 85.23 & \cellcolor{cvprblue!15} 85.78 & \cellcolor{cvprblue!15} 84.47 & \cellcolor{cvprblue!15} 85.12 
 & \cellcolor{cvprblue!15} 88.00 & \cellcolor{cvprblue!15} 97.34 & \cellcolor{cvprblue!15} 78.13 & \cellcolor{cvprblue!15} 86.69 \\
 &  & M3ID  & 86.33 & 85.30 & 87.80 & 86.53 & 85.00 & 84.72 & 85.40 & 85.06 & 86.03 & \underline{97.87} & 73.67 & 84.06 \\
 &  & M3ID + VDC  
 & \cellcolor{cvprblue!15} 86.73 & \cellcolor{cvprblue!15} 85.83 & \cellcolor{cvprblue!15} 88.00 & \cellcolor{cvprblue!15} 86.90 
 & \cellcolor{cvprblue!15} 85.53 & \cellcolor{cvprblue!15} 85.34 & \cellcolor{cvprblue!15} 85.80 & \cellcolor{cvprblue!15} 85.57 
 & \cellcolor{cvprblue!15} 86.50 & \cellcolor{cvprblue!15} \textbf{97.90} & \cellcolor{cvprblue!15} 74.60 & \cellcolor{cvprblue!15} 84.68 \\
 &  & ONLY  & \underline{89.57} & \underline{90.68} & \underline{88.20} & \underline{89.42} & \underline{86.13} & \underline{86.04} & \underline{86.27} & \underline{86.15} & \underline{89.63} & 95.70 & \underline{83.00} & \underline{88.90} \\
 &  & ONLY + VDC  
 & \cellcolor{cvprblue!15} \textbf{89.83} & \cellcolor{cvprblue!15} \textbf{90.73} & \cellcolor{cvprblue!15} \textbf{88.73} & \cellcolor{cvprblue!15} \textbf{89.72} 
 & \cellcolor{cvprblue!15} \textbf{86.50} & \cellcolor{cvprblue!15} \textbf{86.14} & \cellcolor{cvprblue!15} \textbf{87.00} & \cellcolor{cvprblue!15} \textbf{86.57} 
 & \cellcolor{cvprblue!15} \textbf{90.07} & \cellcolor{cvprblue!15} 95.74 & \cellcolor{cvprblue!15} \textbf{83.87} & \cellcolor{cvprblue!15} \textbf{89.41} \\

\arrayrulecolor{gray}\cmidrule(lr){2-15}
 & \multirow{8}{*}{Popular}
 &  Vanilla  & 81.33 & 78.14 & 87.00 & 82.33 & 76.00 & 72.11 & 84.80 & 77.94 & 84.50 & 94.73 & 73.07 & 82.50 \\
 &  & Vanilla + VDC  
 & \cellcolor{cvprblue!15} 81.80 & \cellcolor{cvprblue!15} 78.77 & \cellcolor{cvprblue!15} 87.07 & \cellcolor{cvprblue!15} 82.71 
 & \cellcolor{cvprblue!15} 76.47 & \cellcolor{cvprblue!15} 72.56 & \cellcolor{cvprblue!15} 85.13 & \cellcolor{cvprblue!15} 78.34 
 & \cellcolor{cvprblue!15} 84.67 & \cellcolor{cvprblue!15} 94.91 & \cellcolor{cvprblue!15} 73.27 & \cellcolor{cvprblue!15} 82.69 \\
 &  & VCD  & 80.80 & 77.11 & 87.60 & 82.02 & 77.20 & 74.00 & 83.87 & 78.62 & 85.83 & 94.02 & 76.53 & 84.38 \\
 &  & VCD + VDC  
 & \cellcolor{cvprblue!15} 81.20 & \cellcolor{cvprblue!15} 77.56 & \cellcolor{cvprblue!15} 87.80 & \cellcolor{cvprblue!15} 82.36 
 & \cellcolor{cvprblue!15} 77.23 & \cellcolor{cvprblue!15} \underline{74.04} & \cellcolor{cvprblue!15} 83.87 & \cellcolor{cvprblue!15} 78.65 
 & \cellcolor{cvprblue!15} 85.90 & \cellcolor{cvprblue!15} 94.18 & \cellcolor{cvprblue!15} 76.53 & \cellcolor{cvprblue!15} 84.44 \\
 &  & M3ID  & 82.30 & 79.10 & 87.80 & 83.22 & 77.23 & 73.41 & 85.40 & 78.95 & 85.43 & \underline{95.94} & 74.00 & 83.55 \\
 &  & M3ID + VDC  
 & \cellcolor{cvprblue!15} 82.97 & \cellcolor{cvprblue!15} 79.77 & \cellcolor{cvprblue!15} \underline{88.33} & \cellcolor{cvprblue!15} 83.83 
 & \cellcolor{cvprblue!15} \underline{77.67} & \cellcolor{cvprblue!15} 73.80 & \cellcolor{cvprblue!15} \underline{85.80} & \cellcolor{cvprblue!15} \underline{79.35} 
 & \cellcolor{cvprblue!15} 85.67 & \cellcolor{cvprblue!15} \textbf{96.12} & \cellcolor{cvprblue!15} 74.33 & \cellcolor{cvprblue!15} 83.83 \\
 &  & ONLY  & \underline{86.10} & \underline{84.64} & 88.20 & \underline{86.39} & 77.50 & 73.40 & \textbf{86.27} & 79.31 & \underline{87.70} & 91.92 & \underline{82.67} & \underline{87.05} \\
 &  & ONLY + VDC  
 & \cellcolor{cvprblue!15} \textbf{86.83} & \cellcolor{cvprblue!15} \textbf{85.39} & \cellcolor{cvprblue!15} \textbf{88.87} & \cellcolor{cvprblue!15} \textbf{87.10} 
 & \cellcolor{cvprblue!15} \textbf{78.40} & \cellcolor{cvprblue!15} \textbf{74.54} & \cellcolor{cvprblue!15} \textbf{86.27} & \cellcolor{cvprblue!15} \textbf{79.98} 
 & \cellcolor{cvprblue!15} \textbf{88.03} & \cellcolor{cvprblue!15} 92.29 & \cellcolor{cvprblue!15} \textbf{83.00} & \cellcolor{cvprblue!15} \textbf{87.40} \\

\arrayrulecolor{gray}\cmidrule(lr){2-15}
 & \multirow{8}{*}{Adversarial}
 &  Vanilla  & 75.87 & 71.18 & 86.93 & 78.27 & 74.17 & 70.04 & 84.47 & 76.58 & 82.53 & 90.80 & 72.40 & 80.56 \\
 &  & Vanilla + VDC  
 & \cellcolor{cvprblue!15} 76.63 & \cellcolor{cvprblue!15} 72.01 & \cellcolor{cvprblue!15} 87.13 & \cellcolor{cvprblue!15} 78.85 
 & \cellcolor{cvprblue!15} 74.97 & \cellcolor{cvprblue!15} 70.96 & \cellcolor{cvprblue!15} 84.53 & \cellcolor{cvprblue!15} 77.15 
 & \cellcolor{cvprblue!15} 83.20 & \cellcolor{cvprblue!15} 91.09 & \cellcolor{cvprblue!15} 73.60 & \cellcolor{cvprblue!15} 81.42 \\
 &  & VCD  & 75.23 & 70.23 & 87.60 & 77.96 & 75.80 & \underline{72.29} & 83.67 & 77.56 & 83.10 & 88.46 & 76.13 & 81.83 \\
 &  & VCD + VDC  
 & \cellcolor{cvprblue!15} 75.67 & \cellcolor{cvprblue!15} 70.52 & \cellcolor{cvprblue!15} \underline{88.20} & \cellcolor{cvprblue!15} 78.38 
 & \cellcolor{cvprblue!15} \underline{76.30} & \cellcolor{cvprblue!15} \textbf{72.82} & \cellcolor{cvprblue!15} 83.93 & \cellcolor{cvprblue!15} 77.98 
 & \cellcolor{cvprblue!15} 83.73 & \cellcolor{cvprblue!15} 88.98 & \cellcolor{cvprblue!15} 77.00 & \cellcolor{cvprblue!15} 82.56 \\
 &  & M3ID  & 76.63 & 71.79 & 87.73 & 78.97 & 75.40 & 71.26 & 85.13 & 77.58 & 83.03 & \underline{91.12} & 73.20 & 81.18 \\
 &  & M3ID + VDC  
 & \cellcolor{cvprblue!15} 76.87 & \cellcolor{cvprblue!15} 72.05 & \cellcolor{cvprblue!15} 87.80 & \cellcolor{cvprblue!15} 79.15 
 & \cellcolor{cvprblue!15} \underline{76.30} & \cellcolor{cvprblue!15} 72.20 & \cellcolor{cvprblue!15} 85.53 & \cellcolor{cvprblue!15} \underline{78.30} 
 & \cellcolor{cvprblue!15} 83.40 & \cellcolor{cvprblue!15} \textbf{91.34} & \cellcolor{cvprblue!15} 73.80 & \cellcolor{cvprblue!15} 81.64 \\
 &  & ONLY  & \underline{79.43} & \underline{75.07} & 88.13 & \underline{81.08} & 75.63 & 71.28 & \underline{85.87} & 77.90 & \underline{83.77} & 85.10 & \underline{81.87} & \underline{83.45} \\
 &  & ONLY + VDC  
 & \cellcolor{cvprblue!15} \textbf{79.67} & \cellcolor{cvprblue!15} \textbf{75.23} & \cellcolor{cvprblue!15} \textbf{88.47} & \cellcolor{cvprblue!15} \textbf{81.31} 
 & \cellcolor{cvprblue!15} \textbf{76.40} & \cellcolor{cvprblue!15} 72.02 & \cellcolor{cvprblue!15} \textbf{86.33} & \cellcolor{cvprblue!15} \textbf{78.53} 
 & \cellcolor{cvprblue!15} \textbf{84.40} & \cellcolor{cvprblue!15} 85.78 & \cellcolor{cvprblue!15} \textbf{82.47} & \cellcolor{cvprblue!15} \textbf{84.09} \\
 \bottomrule
    \end{tabular}
    }

\end{table*}

%% file: tables/chair.tex
\begin{table*}[t]
    \vspace{-10pt}
    \caption{{Results on CHAIR.}  $\downarrow$ denotes lower is better. * denotes results reported in prior works~\cite{dola_deco,only_lack_visual,opera_lack_visual}.   -- denotes unavailable results.}
    \vspace{-18pt}
    \label{tab:CHAIR}
    
    \begin{center}
    \begin{small}
    \setlength{\tabcolsep}{2pt}
    \resizebox{\textwidth}{!}{
    \begin{tabular}{lcccccccccccc}
        \toprule
          \multirow{3}{*}[-0.5ex]{{Method}}  &  \multicolumn{4}{c}{{LLaVA-1.5}} & \multicolumn{4}{c}{{InstructBLIP}} & \multicolumn{4}{c}{{Qwen-VL}} \\
        \cmidrule(lr){2-5}\cmidrule(lr){6-9}\cmidrule(lr){10-13}
             & \multicolumn{2}{c}{Max Token 64} & \multicolumn{2}{c}{Max Token 128} & \multicolumn{2}{c}{Max Token 64} & \multicolumn{2}{c}{Max Token 128} & \multicolumn{2}{c}{Max Token 64} & \multicolumn{2}{c}{Max Token 128} \\
         \cmidrule(lr){2-3}\cmidrule(lr){4-5}\cmidrule(lr){6-7}\cmidrule(lr){8-9}\cmidrule(lr){10-11}\cmidrule(lr){12-13}
             & CHAIR$_S$ $\downarrow$ & CHAIR$_I$ $\downarrow$ & CHAIR$_S$ $\downarrow$ & CHAIR$_I$ $\downarrow$ & CHAIR$_S$ $\downarrow$ & CHAIR$_I$ $\downarrow$ & CHAIR$_S$ $\downarrow$ & CHAIR$_I$ $\downarrow$ & CHAIR$_S$ $\downarrow$ & CHAIR$_I$ $\downarrow$ & CHAIR$_S$ $\downarrow$ & CHAIR$_I$ $\downarrow$ \\
         \midrule

Vanilla & 26.5 & 9.4 & 55.1 & 16.4 & 31.5 & 11.4 & 57.4 & 17.6 & 33.8 & 12.9 & 52.1 & 16.7 \\
Vanilla + VDC  
 & \cellcolor{cvprblue!15} 15.6 & \cellcolor{cvprblue!15} 6.2 & \cellcolor{cvprblue!15} 33.6 & \cellcolor{cvprblue!15} 9.8 
 & \cellcolor{cvprblue!15} \underline{21.4} & \cellcolor{cvprblue!15} \underline{8.1} & \cellcolor{cvprblue!15} 38.2 & \cellcolor{cvprblue!15} 9.1 
 & \cellcolor{cvprblue!15} 19.4 & \cellcolor{cvprblue!15} 8.4 & \cellcolor{cvprblue!15} \underline{34.2} & \cellcolor{cvprblue!15} \underline{9.1} \\
VCD & 24.8 & 8.0 & 54.4 & 16.6 & 30.0 & 10.1 & 60.7 & 18.0 & 33.3 & 13.1 & 50.4 & 17.2 \\
VCD + VDC  
 & \cellcolor{cvprblue!15} 19.3 & \cellcolor{cvprblue!15} 7.1 & \cellcolor{cvprblue!15} 32.9 & \cellcolor{cvprblue!15} \underline{9.2} 
 & \cellcolor{cvprblue!15} 24.2 & \cellcolor{cvprblue!15} 8.6 & \cellcolor{cvprblue!15} 38.6 & \cellcolor{cvprblue!15} 9.7 
 & \cellcolor{cvprblue!15} \underline{19.1} & \cellcolor{cvprblue!15} 8.8 & \cellcolor{cvprblue!15} 34.6 & \cellcolor{cvprblue!15} 9.6 \\
M3ID & 21.4 & 6.4 & 56.6 & 15.8 & 31.1 & 10.5 & 62.3 & 18.2 & 32.3 & 11.9 & 49.8 & 17.4 \\
M3ID + VDC  
 & \cellcolor{cvprblue!15} \underline{15.3} & \cellcolor{cvprblue!15} \underline{6.1} & \cellcolor{cvprblue!15} \underline{33.8} & \cellcolor{cvprblue!15} 9.6 
 & \cellcolor{cvprblue!15} \underline{21.6} & \cellcolor{cvprblue!15} 9.2 & \cellcolor{cvprblue!15} \underline{33.9} & \cellcolor{cvprblue!15} \underline{8.9} 
 & \cellcolor{cvprblue!15} 19.2 & \cellcolor{cvprblue!15} \underline{8.1} & \cellcolor{cvprblue!15} 34.2 & \cellcolor{cvprblue!15} 9.7 \\
ONLY & 20.1 & 6.3 & 51.2 & 14.9 & 23.9 & 8.3 & 52.5 & 15.7 & 27.7 & 8.6 & 48.1 & 14.4 \\
ONLY + VDC  
 & \cellcolor{cvprblue!15} \textbf{13.6} & \cellcolor{cvprblue!15} \textbf{5.6} & \cellcolor{cvprblue!15} \textbf{31.4} & \cellcolor{cvprblue!15} \textbf{8.9} 
 & \cellcolor{cvprblue!15} \textbf{17.8} & \cellcolor{cvprblue!15} \textbf{6.6} & \cellcolor{cvprblue!15} \textbf{31.6} & \cellcolor{cvprblue!15} \textbf{7.3} 
 & \cellcolor{cvprblue!15} \textbf{17.8} & \cellcolor{cvprblue!15} \textbf{7.4} & \cellcolor{cvprblue!15} \textbf{32.6} & \cellcolor{cvprblue!15} \textbf{8.7} \\
DOLA$^*$ & -- & --  & 47.8 &  13.8 & -- & -- &  48.4 &  15.9 & -- & -- &  46.8 &  12.9 \\
DeCo$^*$ & -- & --  & 37.8 &  11.1 & -- & -- &  41.2 &  12.4 & -- & -- &  42.2 &  10.4 \\
{OPERA}$^*$ & -- & --  & 44.6 &  12.8 & -- & -- &  46.4 &  14.2 & -- & -- &  -- & --  \\
{Woodpecker}$^*$ & 24.9 & 7.5  & 57.6 & 16.7 &  31.2 & 10.8 & 60.8 & 17.6 & 31.1 & 12.3 & 51.8 & 16.3 \\
{HALC}$^*$ & 21.7 & 7.1  & 51.0 & 14.8 &  24.5 & 8.0 & 53.8 & 15.7 & 28.2 & 9.1 & 49.6 & 15.4 \\
        \bottomrule
    \end{tabular}
    }
\vspace{-12pt}
    \end{small}
    \end{center}
\end{table*}

%% file: tables/mme.tex
\begin{table}
    \vspace{-10pt}
    \caption{Results on MME. $\uparrow$ indicates higher is better.}
    \vspace{-10pt}
    \label{tab:MME}
\footnotesize
\centering
        \resizebox{1\linewidth}{!}{
        \begin{tabular}{lccccc}
            \toprule
                \multirow{2}{*}[-0.5ex]{{Method}} &
                \multicolumn{2}{c}{{Object-level}} &
                \multicolumn{2}{c}{{Attribute-level}} &
                \multirow{2}{*}[-0.5ex]{{MME Score $\uparrow$}} \\
                \cmidrule(lr){2-3}
                \cmidrule(lr){4-5}
                & Existence $\uparrow$ & Count $\uparrow$ & Position $\uparrow$ & Color $\uparrow$ & \\
                \midrule
             Vanilla & 185.00 & 126.67 & 128.33 & 148.33 & 588.33 \\
             Vanilla + VDC  
             & \cellcolor{cvprblue!15} 187.00 
             & \cellcolor{cvprblue!15} 128.67 
             & \cellcolor{cvprblue!15} 130.33 
             & \cellcolor{cvprblue!15} 150.67 
             & \cellcolor{cvprblue!15} 596.67 \\
             VCD & 185.00 & 136.67 & 128.33 & {158.33} & 608.33 \\
             VCD + VDC  
             & \cellcolor{cvprblue!15} 186.00 
             & \cellcolor{cvprblue!15} 138.33 
             & \cellcolor{cvprblue!15} 131.67 
             & \cellcolor{cvprblue!15} \underline{159.67} 
             & \cellcolor{cvprblue!15} 615.67 \\
             M3ID & \underline{190.00} & 136.67 & 128.33 & {158.33} & 613.33 \\
             M3ID + VDC  
             & \cellcolor{cvprblue!15} \textbf{191.00} 
             & \cellcolor{cvprblue!15} 138.33 
             & \cellcolor{cvprblue!15} 130.67 
             & \cellcolor{cvprblue!15} \textbf{160.33} 
             & \cellcolor{cvprblue!15} \underline{620.33} \\
             ONLY  & \underline{190.00} & \underline{143.33} & \underline{133.33} & 148.33 & 614.99 \\
             ONLY + VDC  
             & \cellcolor{cvprblue!15} \textbf{191.00} 
             & \cellcolor{cvprblue!15} \textbf{144.67} 
             & \cellcolor{cvprblue!15} \textbf{134.67} 
             & \cellcolor{cvprblue!15} 150.33 
             & \cellcolor{cvprblue!15} \textbf{620.67} \\
            \bottomrule
        \end{tabular}
        }

\end{table}

%% file: tables/ablation.tex
\begin{table}[t]
\vspace{-10pt}
\caption{Ablation results on validation sources. }
\vspace{-10pt}
\label{tab:vdc_ablation_validation}
\footnotesize
\centering
\resizebox{0.95\linewidth}{!}{
\begin{tabular}{lccc}
\toprule
\textbf{Validation Sources} & \textbf{CHAIR$_S$} $\downarrow$ & \textbf{CHAIR$_I$} $\downarrow$ & \textbf{Recall} $\uparrow$ \\
\midrule
Baseline        & 55.1 & 16.4 & \textbf{70.6} \\
Layer-only       & 47.1 & 11.3 &  64.2 \\
Attn+FFN         & \textbf{33.6} & \textbf{9.8} & \underline{69.9} \\
Attn+FFN+Layer   & \underline{44.2} & \underline{10.6} & 68.4  \\
\bottomrule
\end{tabular}
}
\end{table}

\begin{table}[t]
\vspace{-5pt}
\caption{Ablation results on correction sources.}
\vspace{-10pt}
\label{tab:vdc_ablation_replacement}
\footnotesize
\centering
\resizebox{0.95\linewidth}{!}{
\begin{tabular}{lccc}
\toprule
\textbf{Correction Sources} & \textbf{CHAIR$_S$} $\downarrow$ & \textbf{CHAIR$_I$} $\downarrow$ & \textbf{Recall} $\uparrow$ \\
\midrule
Baseline        & 55.1 & 16.4 & \textbf{70.6} \\
Layer-only       & 46.3 & 11.6 &  64.2 \\
Attn+FFN   & \textbf{26.2} & \textbf{8.6} & 44.6  \\
Attn+FFN+Layer        & \underline{33.6} & \underline{9.8} & \underline{69.9} \\

\bottomrule
\end{tabular}
}

\end{table}

\begin{table}[t]
\vspace{-5pt}
\caption{Ablation on skipping early layers in VDC. }
\vspace{-10pt}
\label{tab:skip_layers_ablation}
\footnotesize
\centering
\resizebox{0.95\linewidth}{!}{
\begin{tabular}{lccc}
\toprule
\textbf{Skipped Early Layers} & \textbf{CHAIR$_S$} $\downarrow$ & \textbf{CHAIR$_I$} $\downarrow$ & \textbf{Recall} $\uparrow$ \\
\midrule
None (Full)       & {33.6} & {9.8} & {69.9} \\
2 layers       & 33.6 & 9.8 &  69.9 \\
10 layers      & 33.6 & 9.8 &  69.9 \\
16 layers      & 33.4 & 10.1 &  69.6 \\
\bottomrule
\end{tabular}
}
\vspace{-8pt}
\end{table}

%% file: sections/5.conclusion.tex
\section{Conclusion}
\label{sec:conclusion}

In this work, we systematically analyze the perception–generation dynamics of LVLMs.
We uncover the \textbf{GATE} (Global, Approach \& Tighten, Explore) pattern in visual perception and the \textbf{SAD} (Subdominant Accumulation to Dominant) pattern in token generation, revealing how hallucinations emerge from subdominant token accumulation.
Based on these insights, we propose \textbf{VDC} (Validated Dominance Correction), a lightweight strategy to reliably detect and mitigate hallucinations.

%% file: sections/appendix.tex
\clearpage
\appendix
\setcounter{page}{1}
\maketitlesupplementary

\section{More Examples}
\label{sec:app_more_examples}

We show multiple examples of both correct and hallucination predictions in Figs.~\ref{fig:example_black_apple},~\ref{fig:example_right_brick},~\ref{fig:example_dog_run_away},~\ref{fig:example_person_sitting},~\ref{fig:example_dog_standing_walk},~\ref{fig:example_woman_behind_ahead},~\ref{fig:example_ffn_right_standing}, and~\ref{fig:examples_right_balck_apple}.

From the perception perspective, the model consistently follows the \textbf{GATE} pattern: it first attends to the overall scene (Global), then gradually approaches the target region (Approach), focuses tightly on the relevant area (Tighten), and finally explores nearby regions (Explore).

From the token generation perspective, sudden shifts in the final one or two layers frequently occur. In hallucinated examples, the final output tokens often never appear as dominant (rank-1) tokens in either the attention or FFN streams. Instead, they repeatedly appear among subdominant tokens (ranks 2–5) and, through accumulation across layers, eventually become dominant. This observation further supports the \textbf{SAD} pattern, where subdominant, partially incorrect tokens gradually overtake previously correct dominant predictions.

Leveraging these insights, we propose \textbf{VDC} to detect and mitigate hallucinations. VDC identifies tokens that fail to receive validation from either visual perception (attention) or internal knowledge (FFN) and replaces them with previously validated dominant tokens, thereby reducing hallucination outputs and improving model reliability.

\section{Evaluation on More Methods and Models}
\label{sec:app_more_models_llavanext_chair}

To further verify the generality and robustness of VDC, we extend our evaluation to a broader set of \textbf{hallucination mitigation methods}, including DoLA~\cite{dola}, OPERA~\cite{opera_lack_visual}, VCD~\cite{vcd_lack_visual}, Woodpecker~\cite{woodpecker},  
LURE~\cite{zhouanalyzing_lure}, HALC~\cite{halc_lack_visual}, CODE~\cite{kim2024code}, EAH~\cite{zhang2024seeing_eah}, and VHR~\cite{he2025cracking_vhr}, as well as diverse \textbf{LVLM architectures}, including \textbf{MiniGPT-4}~\cite{minigpt4}, \textbf{mPLUG-Owl2}~\cite{mplugowl2}, and the more recent and stronger \textbf{LLaVA-NeXT}~\cite{llavanext}.  
As shown in Tab~\ref{tab:abla_models_llavanet} and~\ref{tab:abla_models_minigpt_mplugowl}, VDC consistently achieves the lowest CHAIRs and CHAIRi scores, demonstrating its effectiveness across model architectures.

\section{Additional Observations}
\label{sec:app_additional_observations}

\begin{figure*}[t]
    \centering
    \includegraphics[width=0.99\linewidth]{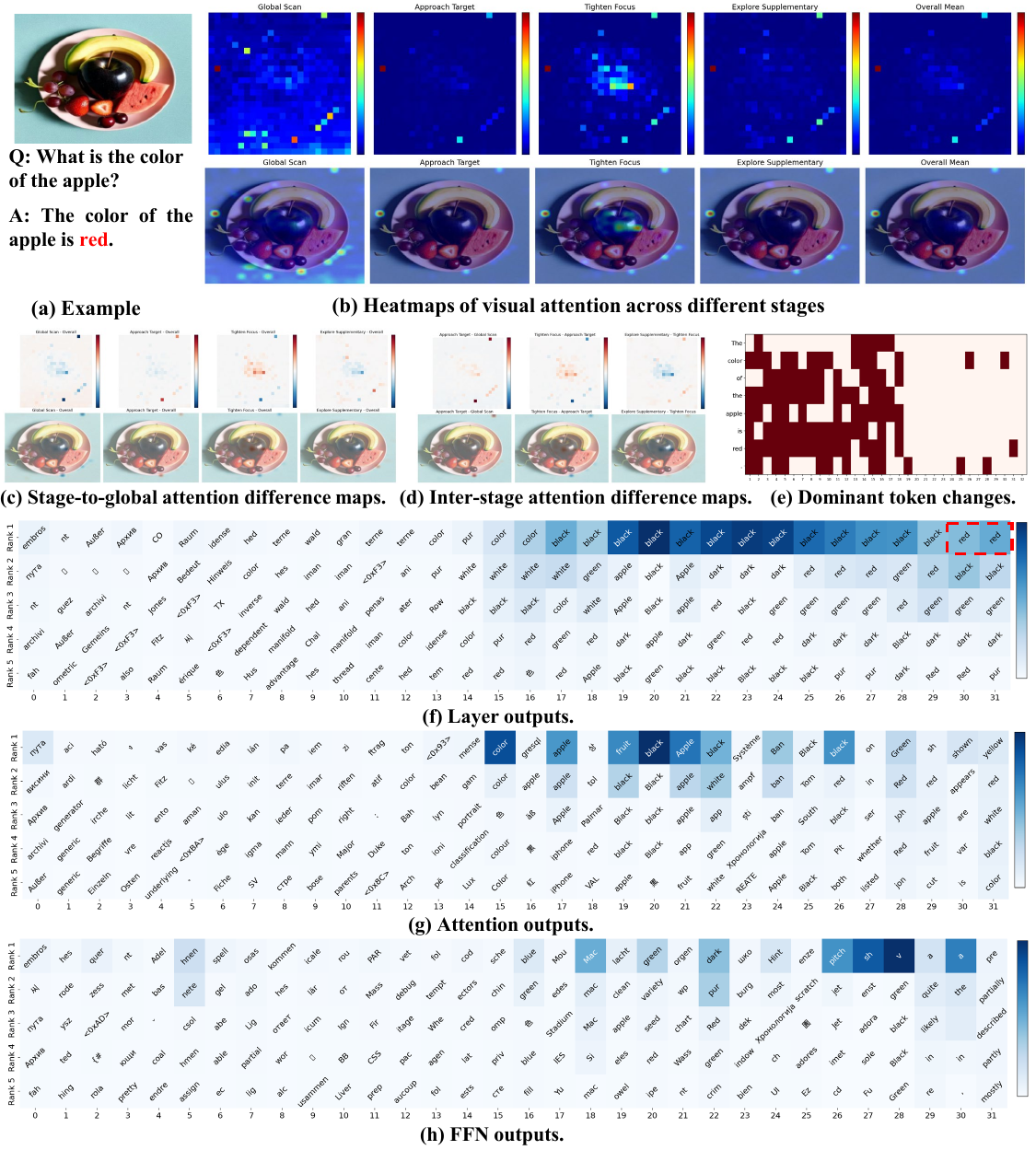}
    \vspace{-10pt}
\caption{
An example of a sudden shift occurring in the last two layers. 
From the perception perspective, during the Tighten phase, the model correctly attends to the target region.  
From the token generation perspective, the model produces the correct token ``black'' in intermediate layers but shifts to the incorrect token ``red'' in the final two layers. 
As shown in panels (f) (g) and (h), the token ``red'' never appears as the dominant (rank-1) token in either the attention or FFN streams, but frequently appears among subdominant tokens (ranks 2--5). 
Through repeated accumulation, these subdominant signals gradually dominate, resulting in the incorrect output and illustrating the \textbf{SAD} pattern.
}

    \vspace{-6pt}
    \label{fig:example_black_apple}
\end{figure*}

\begin{figure*}[t]
    \centering
    \includegraphics[width=0.99\linewidth]{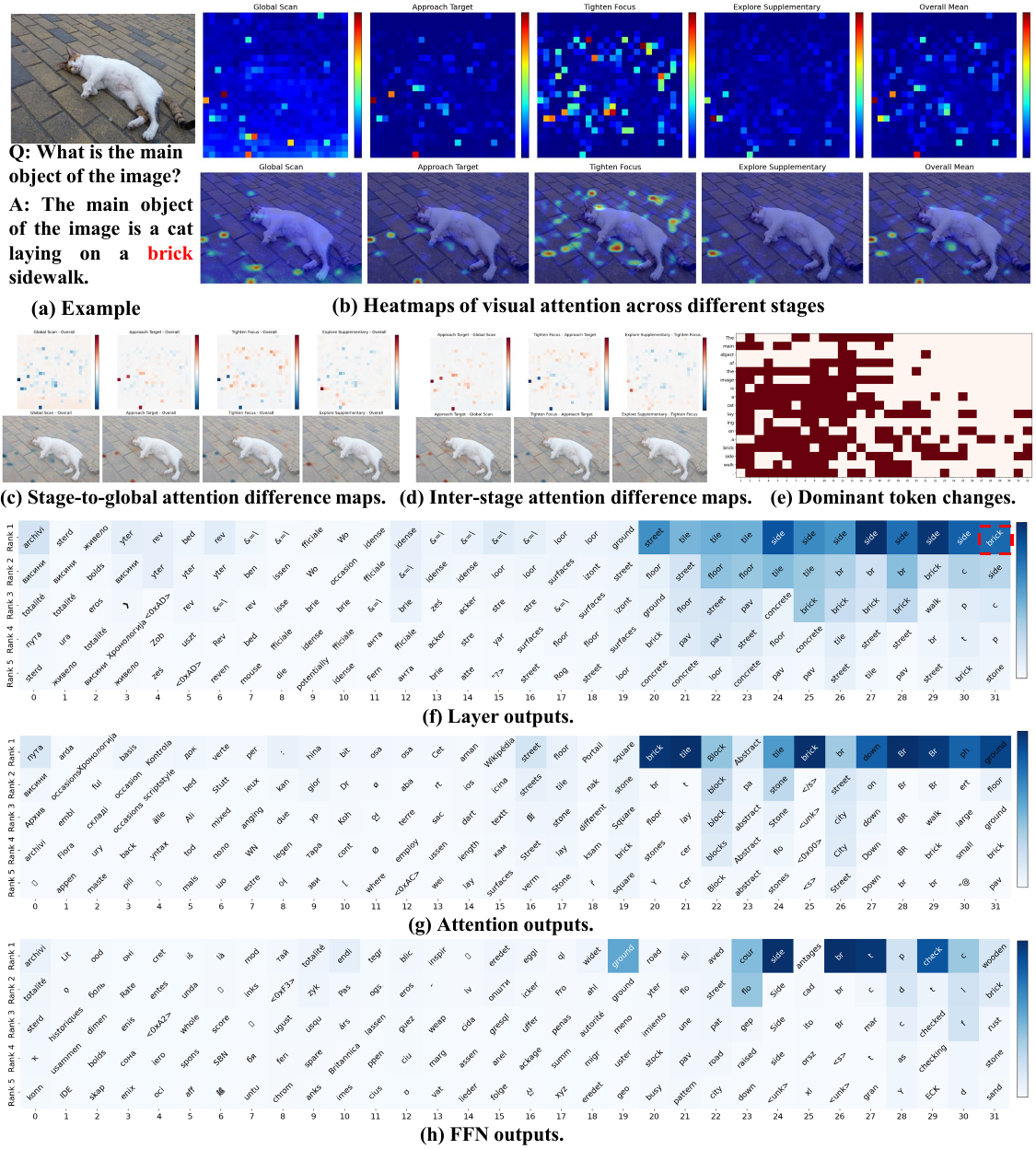}
    \vspace{-10pt}
\caption{
A right example of a sudden shift in the final layer. 
During token generation, the model outputs the token ``brick'' at the last layer, which has already appeared as a dominant token in the attention stream, thus requiring no correction. Other tokens such as ``side'' and ``tile'' are also correct, but ``brick'' is more challenging and demonstrates higher precision. This indicates that not all late-layer sudden shifts are erroneous, highlighting the necessity of the last several layers. Therefore, we term the last stage \textit{Explore Supplementary}.
From the perception perspective, the model first attends to the overall scene (Global), then gradually approaches the brick regions near the cat (Approach), focuses tightly on the surrounding bricks (Tighten), and finally explores nearby areas (Explore).  
(e) further illustrates that the dynamics of sudden shifts vary with token difficulty: simple tokens such as ``the,'' ``main,'' and ``object'' stabilize quickly, whereas more complex tokens like ``laying,'' ``brick,'' and ``side'' continue to fluctuate across layers. 
}

    \vspace{-6pt}
    \label{fig:example_right_brick}
\end{figure*}

\begin{figure*}[t]
    \centering
    \includegraphics[width=0.99\linewidth]{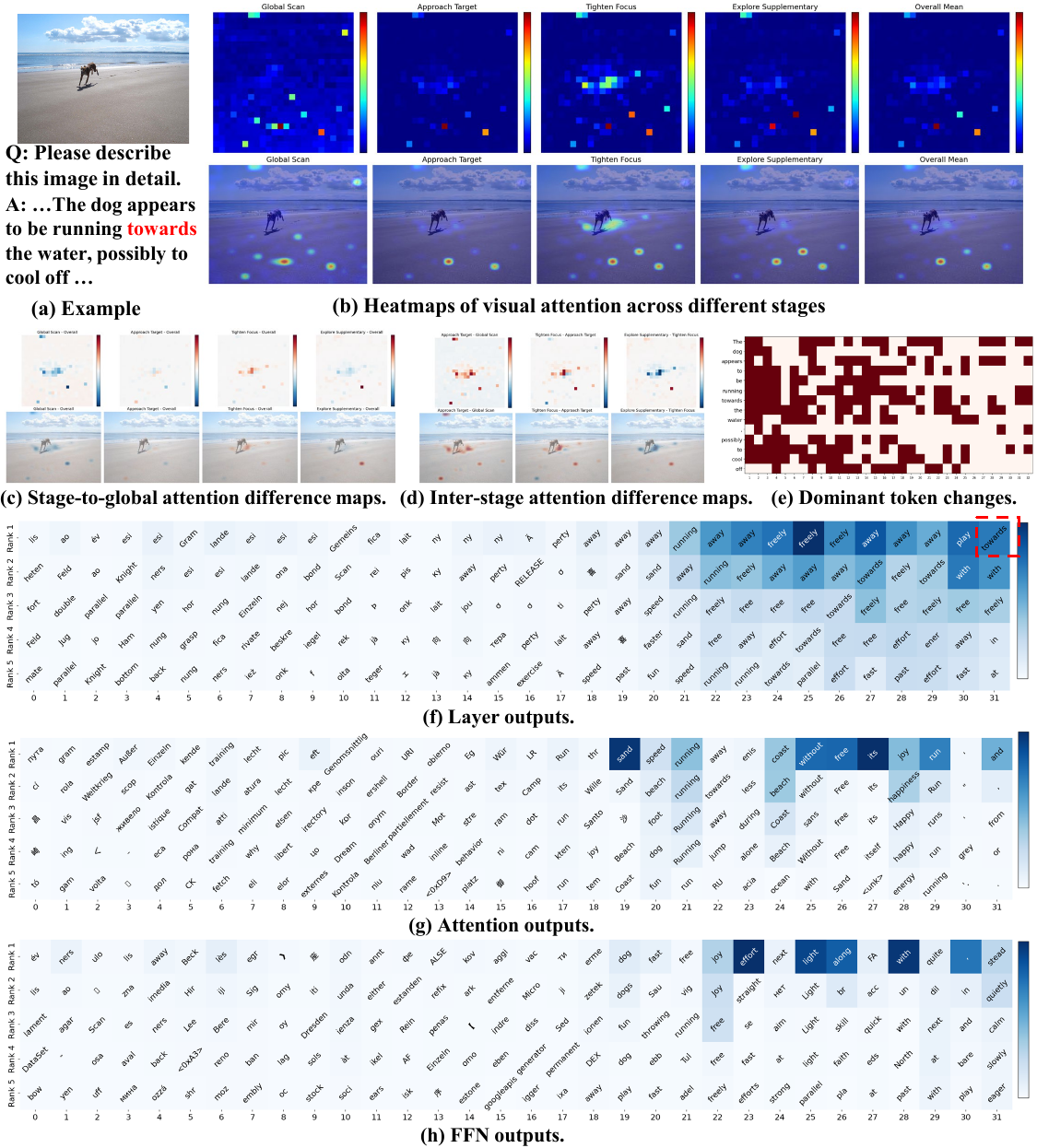}
    \vspace{-10pt}
\caption{
A hallucination example of a sudden shift in the final layer. 
From the perception perspective, during the Tighten phase, the model correctly attends to the target region.  
From the token generation perspective, the model produces the correct token ``away'' in intermediate layers but shifts to the incorrect token ``towards'' in the final layer. 
As shown in panels (f) (g) and (h), the token ``towards'' never appears as the dominant (rank-1) token in either the attention or FFN streams, but frequently appears among subdominant tokens (ranks 2--5). 
Through repeated accumulation, these subdominant signals gradually dominate, resulting in the incorrect output and illustrating the \textbf{SAD} pattern.
}

    \vspace{-6pt}
    \label{fig:example_dog_run_away}
\end{figure*}

\begin{figure*}[t]
    \centering
    \includegraphics[width=0.99\linewidth]{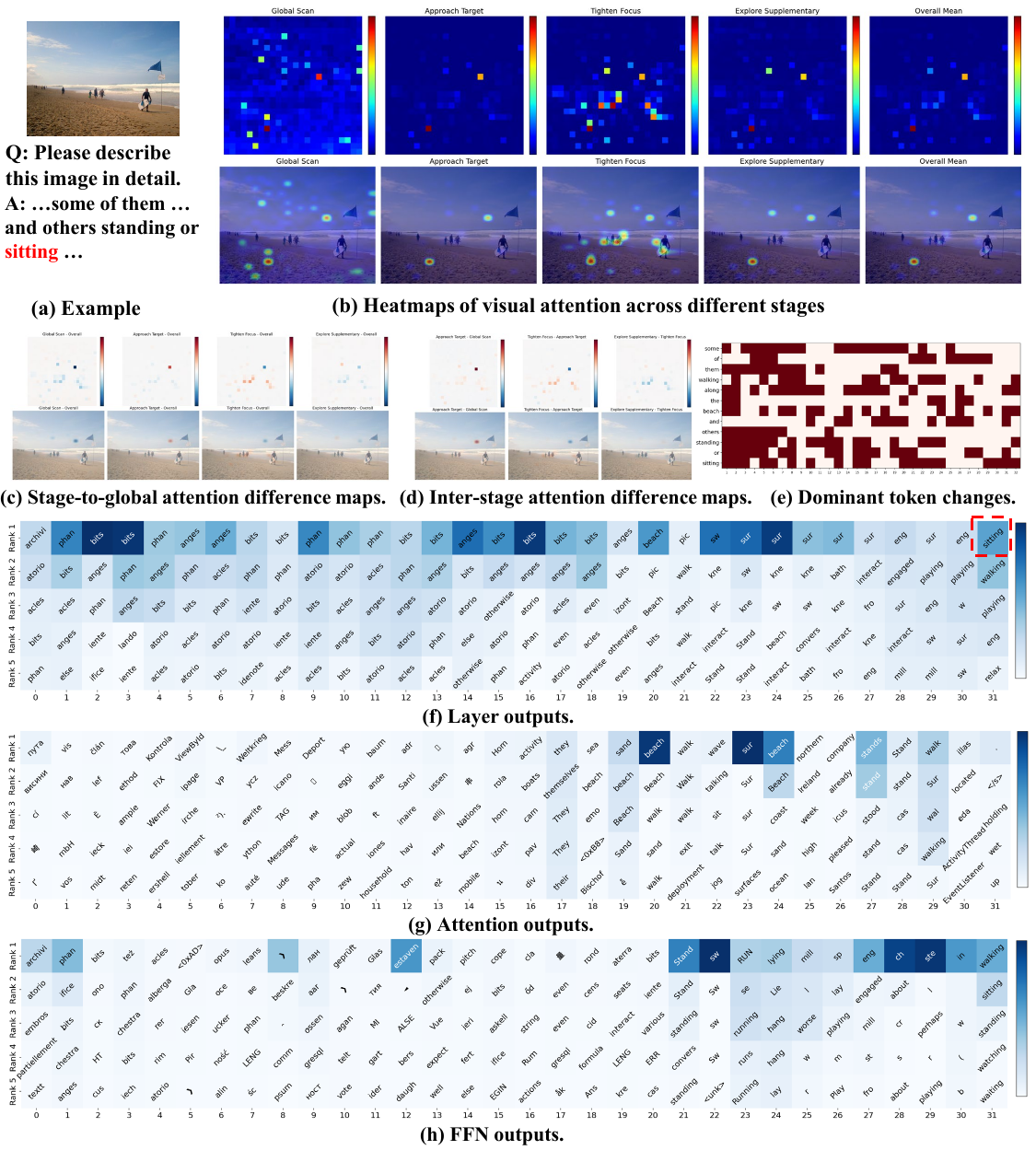}
    \vspace{-10pt}
\caption{
A hallucination example of a sudden shift in the final layer. 
From the perception perspective, during the Tighten phase, the model correctly attends to the target region.  
From the token generation perspective, the model produces the correct token ``sur'' in intermediate layers but shifts to the incorrect token ``sitting'' in the final layer. 
As shown in panels (f) (g) and (h), the token ``sitting'' never appears as the dominant (rank-1) token in either the attention or FFN streams, but frequently appears among subdominant tokens (ranks 2--5). 
Through repeated accumulation, these subdominant signals gradually dominate, resulting in the incorrect output and illustrating the \textbf{SAD} pattern.
}

    \vspace{-6pt}
    \label{fig:example_person_sitting}
\end{figure*}

\begin{figure*}[t]
    \centering
    \includegraphics[width=0.99\linewidth]{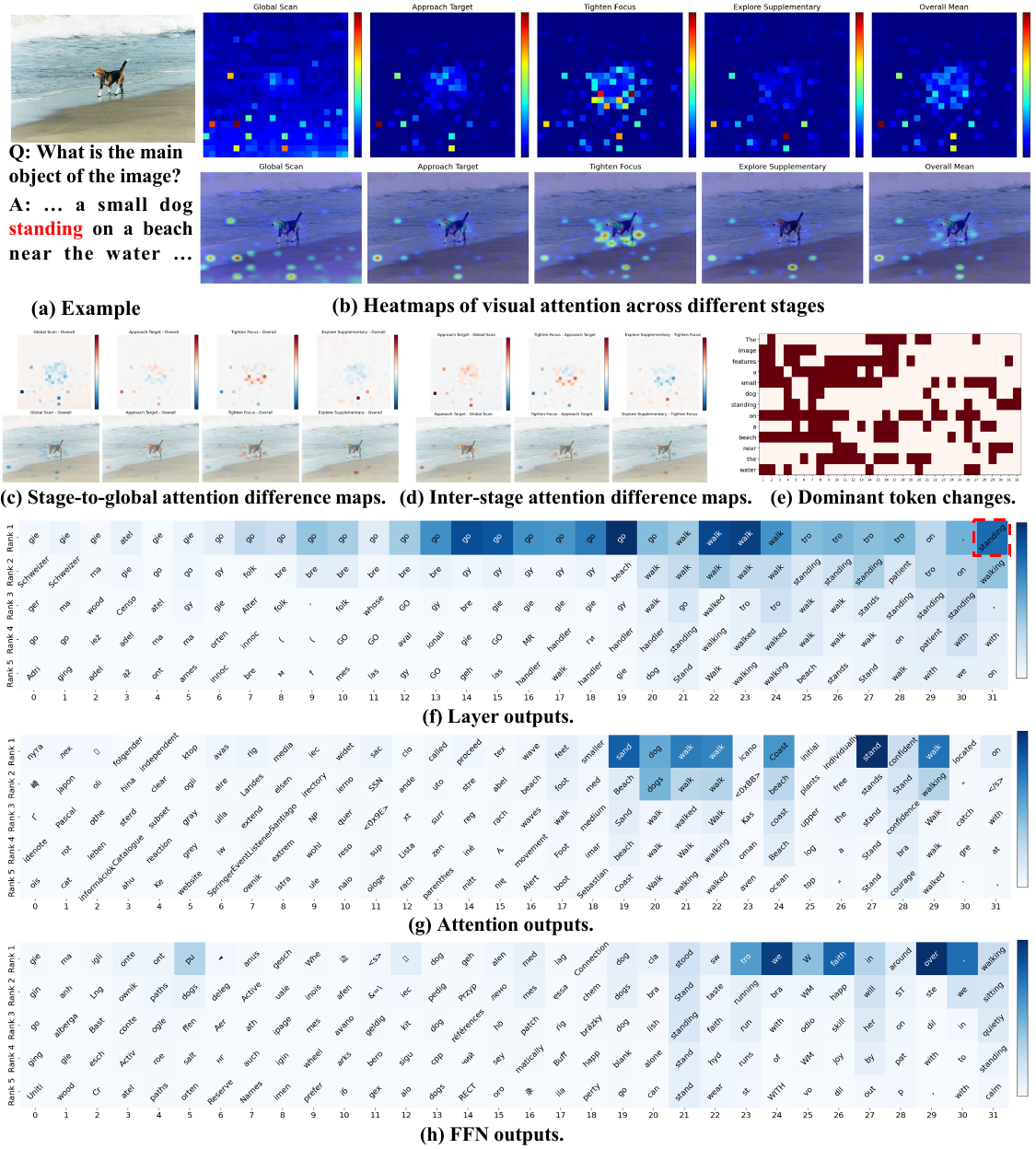}
    \vspace{-10pt}
\caption{
An hallucination example of a sudden shift in the final layer. 
From the perception perspective, during the Tighten phase, the model correctly attends to the target region.  
From the token generation perspective, the model produces the correct tokens (``go'', ``walk'', ``tro'')  in intermediate layers but shifts to the incorrect token ``standing'' in the final layer. 
As shown in panels (f) (g) and (h), the token ``standing'' never appears as the dominant (rank-1) token in either the attention or FFN streams, but frequently appears among subdominant tokens (ranks 2--5). 
Through repeated accumulation, these subdominant signals gradually dominate, resulting in the incorrect output and illustrating the \textbf{SAD} pattern.
}

    \vspace{-6pt}
    \label{fig:example_dog_standing_walk}
\end{figure*}

\begin{figure*}[t]
    \centering
    \includegraphics[width=0.99\linewidth]{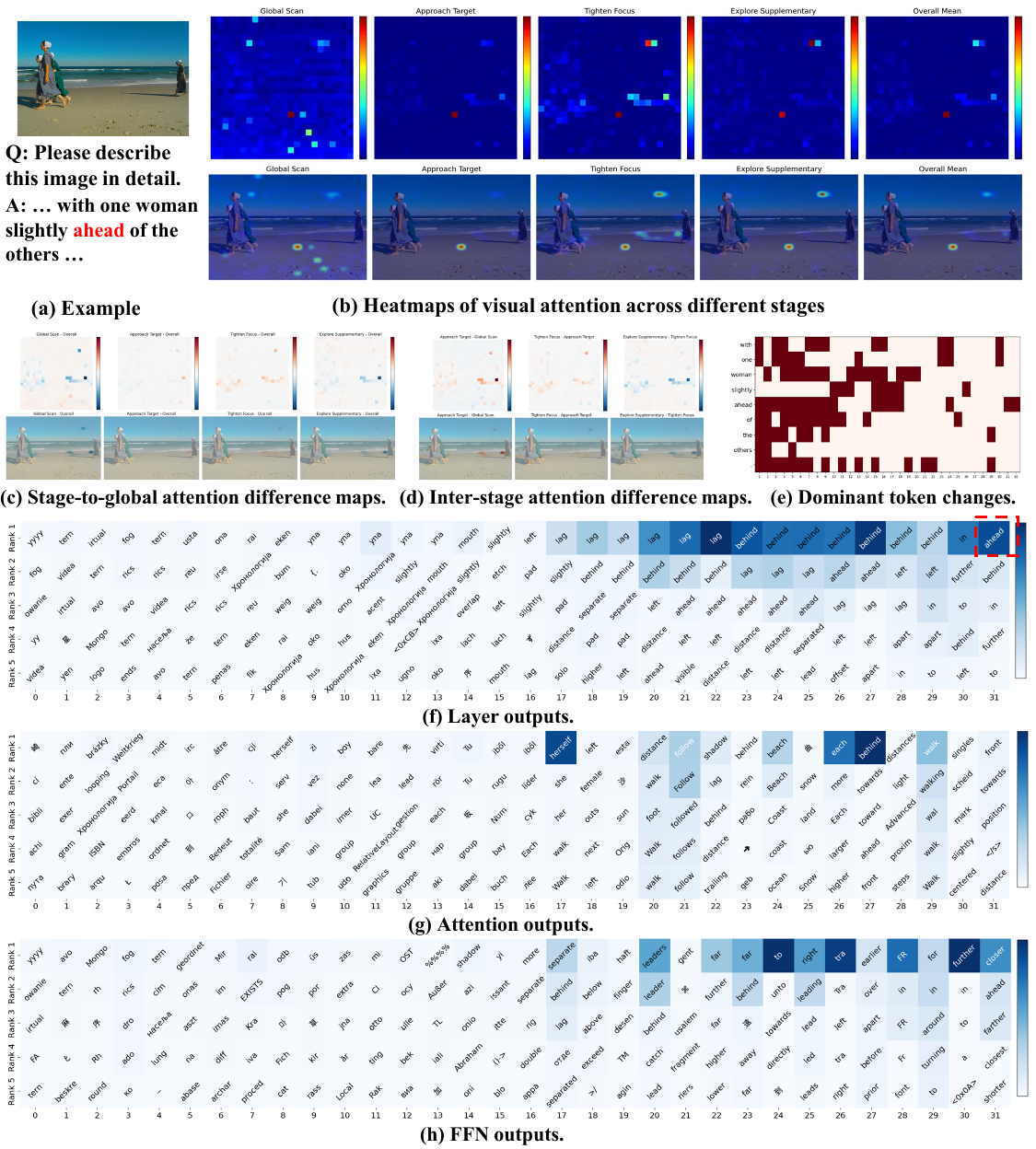}
    \vspace{-10pt}
\caption{
A hallucination example of a sudden shift in the final layer. 
From the perception perspective, during the Tighten phase, the model correctly attends to the target region.  
From the token generation perspective, the model produces the correct token ``behind'' in intermediate layers but shifts to the incorrect token ``ahead'' in the final layer. 
As shown in panels (f) (g) and (h), the token ``ahead'' never appears as the dominant (rank-1) token in either the attention or FFN streams, but frequently appears among subdominant tokens (ranks 2--5). 
Through repeated accumulation, these subdominant signals gradually dominate, resulting in the incorrect output and illustrating the \textbf{SAD} pattern.
}
    \vspace{-6pt}
    \label{fig:example_woman_behind_ahead}
\end{figure*}

\begin{figure*}[t]
    \centering
    \includegraphics[width=0.99\linewidth]{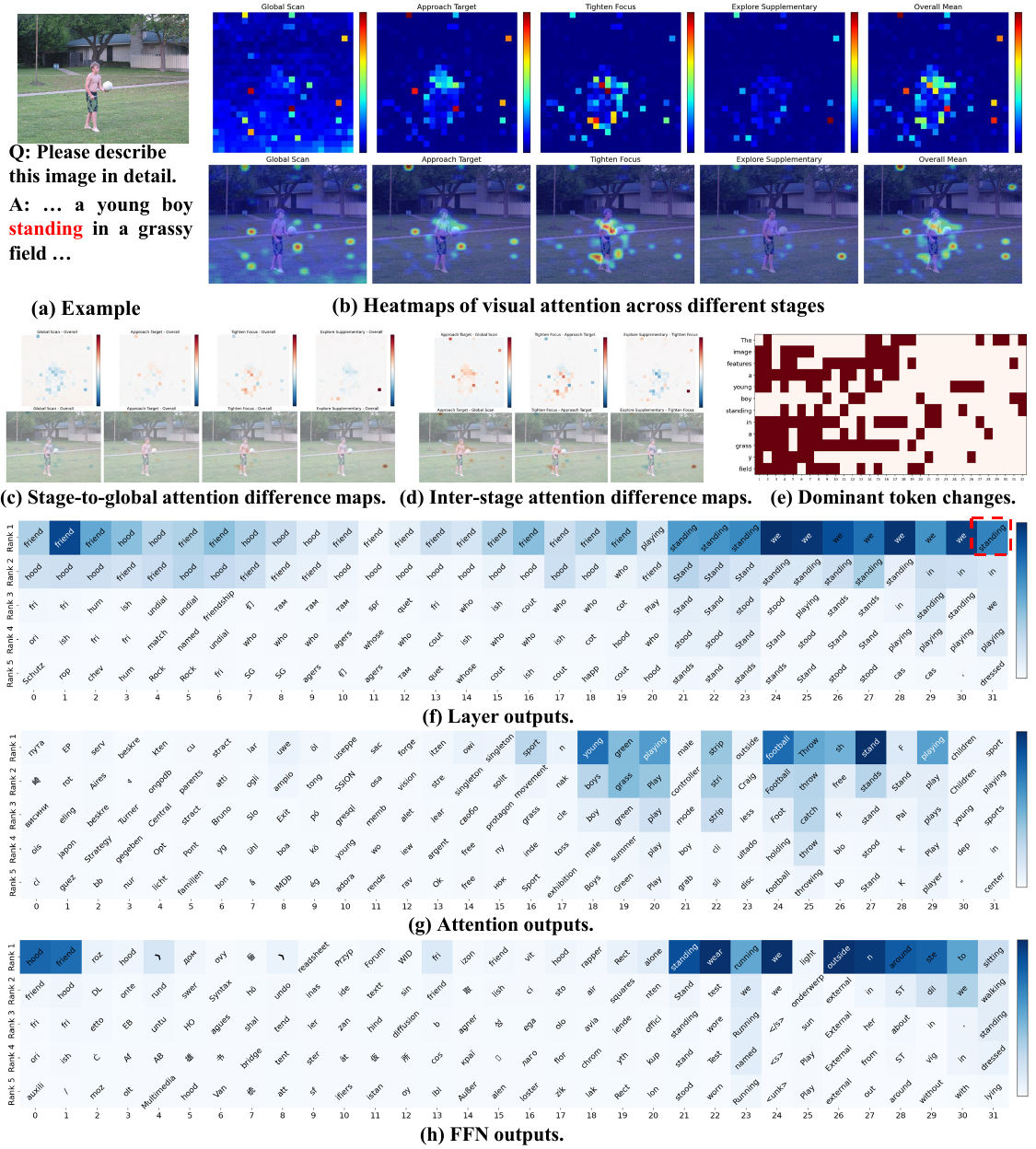}
    \vspace{-10pt}
\caption{
A right example illustrating the role of the FFN in token generation. 
The model outputs the token ``standing'' at the final layer. 
While the attention stream primarily predicts ``stand,'' the FFN stream captures the correct grammatical form ``standing,'' reflecting its role in enhancing syntax-related information. 
The overall visual attention still follows the GATE pattern.
}

    \vspace{-6pt}
    \label{fig:example_ffn_right_standing}
\end{figure*}

\begin{figure*}[t]
    \centering
    \includegraphics[width=0.99\linewidth]{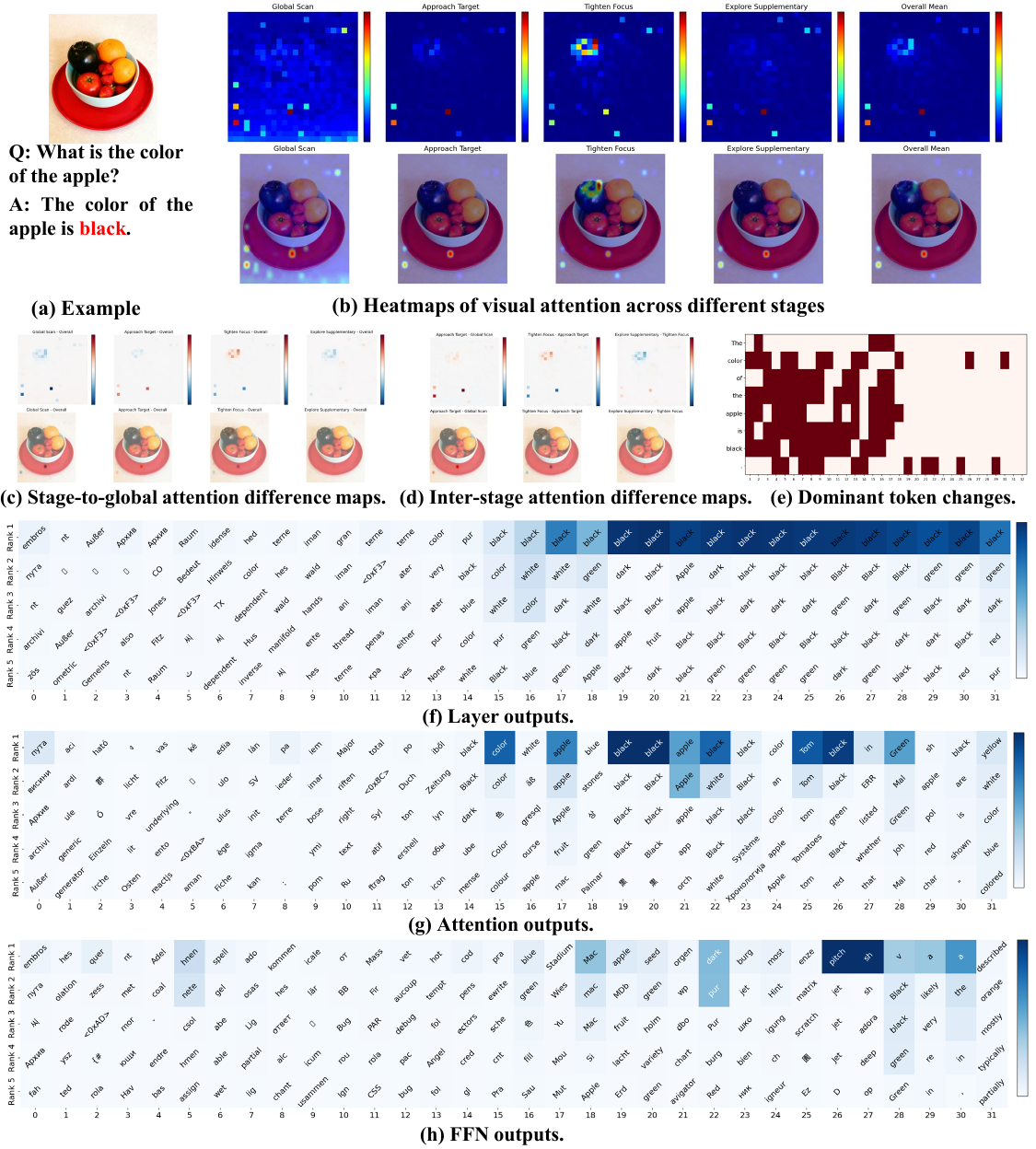}
    \vspace{-10pt}
\caption{
An example where the model correctly predicts the color of the apple as ``black,'' despite strong surrounding distractions. 
This demonstrates that the failure in Figure~\ref{fig:example_black_apple} is not due to the model's inability to learn the ``black apple'' association, but rather a result of internal mechanism issues.  
The visual attention follows the GATE pattern: the model first attends to the overall scene (Global), then gradually approaches the apple region (Approach), tightly focuses on the apple itself (Tighten), and finally explores nearby areas (Explore).  
During generation, the token ``black'' remains consistently predicted across layers and appears as a dominant token in the attention stream.  
According to VDC, this indicates that ``black'' has been validated and requires no correction.
}

    \vspace{-6pt}
    \label{fig:examples_right_balck_apple}
\end{figure*}

\noindent\textbf{Token Dynamics and Sample Difficulty.}
Beyond the GATE (perception) and SAD (generation) phenomena, we observe clear layer-wise dynamics in dominant token evolution (Figs.~\ref{fig:example_black_apple}--\ref{fig:examples_right_balck_apple}, panel e). Early layers show frequent token fluctuations, whereas later layers reveal a natural separation between easy and difficult tokens. Simple or common tokens stabilize quickly, reflecting straightforward predictions, while complex or semantically rich tokens continue to fluctuate, indicating higher uncertainty and stronger reasoning requirements.  This stabilization pattern provides actionable signals for both efficiency and sample difficulty estimation. Tokens that stabilize early can be safely skipped in later computation, enabling potential acceleration via adaptive early-exit or selective inference~\cite{schuster2022confident_skip_layers,kao2020bert_skip_layers,elbayad2019depth_skip_layers}. Persistently fluctuating tokens highlight harder instances, which can guide dataset curation, focused training, or targeted evaluation~\cite{wang2024greats_data_curation_select,bansal2025honeybee_data_curation_select,brown2025benchmark_data_curation_select}. Beyond hallucination analysis, these insights may also benefit other multimodal tasks involving token-level uncertainty, such as curriculum learning, active learning, or adaptive inference strategies.

\noindent\textbf{Submodule Functional Roles.}
Analysis of attention (panel g) and FFN outputs (panel h) further suggests distinct functional roles~\cite{yu2024neuron_ffn_vs_attention,wei2024building_ffn_vs_attention,zhou2024unibias_ffn_vs_attention,kobayashianalyzing_ffn_vs_attention,behrens2024understanding_ffn_vs_attention,kobayashi2021incorporating_ffn_vs_attention}. FFN outputs often contain incomplete or auxiliary tokens, such as grammatical connectors or contextual fillers, whereas attention outputs produce semantically grounded and visually aligned tokens, including nouns, verbs, and adjectives. This pattern indicates that attention primarily supports content grounding, while FFN contributes to linguistic refinement and contextual coherence, potentially reflecting a complementary division of labor in LVLM decoding. Further research is needed to validate and generalize this observation.

Overall, our unified perception–generation analysis reveals internal dynamic processes that static attention or attribution metrics fail to capture, offering deeper mechanistic understanding and practical guidance for building more interpretable and reliable LVLMs.

\begin{table}[t]
\vspace{-6pt}
\caption{Results of MiniGPT-4 and mPLUG-Owl2 on CHAIR with a maximum token length of 64, following the HALC setting.} 
 \vspace{-10pt}
\label{tab:abla_models_minigpt_mplugowl}
\centering
\resizebox{\linewidth}{!}{
\begin{tabular}{lcc|cc}
\toprule
\multirow{2}{*}{Method} & \multicolumn{2}{c|}{MiniGPT-4} & \multicolumn{2}{c}{mPLUG-Owl2} \\
\cmidrule(lr){2-3} \cmidrule(lr){4-5}
& CHAIRs $\downarrow$ & CHAIRi $\downarrow$ & CHAIRs $\downarrow$ & CHAIRi $\downarrow$ \\
\midrule
DoLA          & 30.87 $\pm$ 5.52 & 11.70 $\pm$ 0.13 & 24.60 $\pm$ 0.24 & 8.73 $\pm$ 0.30 \\
OPERA         & 30.00 $\pm$ 4.43 & 11.67 $\pm$ 0.22 & 22.13 $\pm$ 0.86 & 7.57 $\pm$ 1.22 \\
VCD           & 30.27 $\pm$ 0.44 & 12.60 $\pm$ 0.45 & 27.27 $\pm$ 7.32 & 9.73 $\pm$ 1.22 \\
Woodpecker    & 28.87 $\pm$ 2.20 & 10.20 $\pm$ 0.85 & 26.33 $\pm$ 1.98 & 8.43 $\pm$ 0.80 \\
LURE          & 27.88 $\pm$ 2.25 & 10.20 $\pm$ 0.85 & 21.27 $\pm$ 0.06 & 7.67 $\pm$ 0.11 \\
HALC & 17.80 $\pm$ 0.03 & 8.10 $\pm$ 0.14 & 17.33 $\pm$ 4.30 & 7.43 $\pm$ 0.11 \\
\textbf{VDC (ours)} & \textbf{14.68} $\pm$ 0.04 & \textbf{7.42} $\pm$ 0.12 & \textbf{15.58} $\pm$ 4.52 & \textbf{6.84} $\pm$ 0.12 \\
\bottomrule
\end{tabular}}

\end{table}

\begin{table}[t]
\vspace{-10pt}
\caption{Results of the newer and stronger LLaVA-NeXT with a maximum token length of 128, following the VHR setting.}
\vspace{-10pt}
\label{tab:abla_models_llavanet}

\centering
\resizebox{0.6\linewidth}{!}{
\begin{tabular}{lcc}
\toprule
Method & CHAIRs $\downarrow$ & CHAIRi $\downarrow$ \\
\midrule
DoLa   & 28.76$\pm$2.58 & 8.12$\pm$0.78 \\
VCD    & 30.80$\pm$2.48 & 8.72$\pm$0.94 \\
CODE   & 27.84$\pm$2.73 & 7.98$\pm$0.92 \\
EAH    & 28.13$\pm$1.13 & {6.62}$\pm$0.49 \\
VHR    & {24.96}$\pm$2.09 & {6.80}$\pm$0.59 \\
\textbf{VDC (ours)} & \textbf{20.84}$\pm$1.62 & \textbf{6.28}$\pm$0.54 \\
\bottomrule
\end{tabular}}
\vspace{-6pt}

\end{table}

